\documentclass{article}

\usepackage{graphicx}
\usepackage{caption}
\usepackage{subcaption}
\usepackage{extarrows}
\usepackage{multirow,amssymb}
\usepackage{amsmath}
\usepackage{booktabs} 
\usepackage{makecell}
\usepackage{enumitem}
\usepackage{pifont}
\usepackage{bbm}
\newtheorem{prop}{Proposition}

\newtheorem{cor}{Corollary}

\newtheorem{lm}{Lemma}

\newtheorem{thm}{Theorem}

\newcommand{\be}{\begin{eqnarray}}
\newcommand{\ee}{\end{eqnarray}}
\newcommand{\benn}{\begin{eqnarray*}}
\newcommand{\eenn}{\end{eqnarray*}}
\def\IR{\rm I \kern-0.20em R}

\newcommand{\bthm}{\begin{thm}}
\newcommand{\ethm}{\end{thm}}

\newcommand{\bcor}{\begin{cor}}
\newcommand{\ecor}{\end{cor}}
\newcommand{\bprop}{\begin{prop}}
\newcommand{\eprop}{\end{prop}}
\newcommand{\blm}{\begin{lm}}
\newcommand{\elm}{\end{lm}}
\newcommand{\beq}{\begin{equation}}
\newcommand{\eeq}{\end{equation}}
\newcommand{\ber}{\begin{eqnarray}}
\newcommand{\eer}{\end{eqnarray}}

\newcommand{\bproof}{\begin{proof}}
\newcommand{\eproof}{\end{proof}}

\newcommand{\bit}{\begin{itemize}}
\newcommand{\eit}{\end{itemize}}
\newcommand{\ben}{\begin{enumerate}}
\newcommand{\een}{\end{enumerate}}
\newcommand{\bdesc}{\begin{description}}
\newcommand{\edesc}{\end{description}}
\newcommand{\beqarrn}{\begin{eqnarray*}}
\newcommand{\eeqarrn}{\end{eqnarray*}}
\newcommand{\bproofof}{\begin{proofof}}
\newcommand{\eproofof}{\end{proofof}}
\newenvironment{rem}{\begin{trivlist}\item[]{\bf
Remark:}\hspace{4mm}}{\end{trivlist}}
\newcommand{\brem}{\begin{rem}}
\newcommand{\erem}{\end{rem}}
\newenvironment{rems}{\begin{trivlist}\item[]{\bf
Remarks}\begin{itemize}}{\end{itemize}\end{trivlist}}
\newcommand{\brems}{\begin{rems}}
\newcommand{\erems}{\end{rems}}
\newtheorem{fact}{Fact}
\newcommand{\bfact}{\begin{fact}}
\newcommand{\efact}{\end{fact}}
\newtheorem{examp}{Example}
\newcommand{\bexamp}{\begin{examp}\rm}
\newcommand{\eexamp}{\end{examp}}
\newtheorem{defn}{Definition}
\newcommand{\bdefn}{\begin{defn}\rm}
\newcommand{\edefn}{\end{defn}}

\newtheorem{alg}{Algorithm}
\newcommand{\balg}{\begin{alg}}
\newcommand{\ealg}{\end{alg}}

\newtheorem{prob}{Problem}
\newcommand{\bprob}{\begin{prob}}
\newcommand{\eprob}{\end{prob}}

\newcommand{\bvtm}{\begin{verbatim}}
\newcommand{\bfig}{\begin{figure}}
\newcommand{\efig}{\end{figure}}
\newcommand{\bcen}{\begin{center}}
\newcommand{\ecen}{\end{center}}

\long\def\comment#1{}

\def \n2{{N_0 \over 2}}

\def \h5{\hspace{0.5in}}

\newcommand{\dff}{\stackrel{\triangle}{=}}

\usepackage{amsthm}

\newtheorem{theorem}{Theorem}

\newtheorem{definition}{Definition}

\newtheorem{lemma}{Lemma}
\newtheorem{proposition}{Proposition}

\graphicspath{{./figures/}}

\newcommand{\indep}{\perp \!\!\! \perp}
\newcommand{\blink}{\vcenter{\hbox{\includegraphics[scale=0.2]{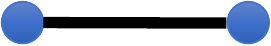}}}}
\newcommand{\rlink}{\vcenter{\hbox{\includegraphics[scale=0.2]{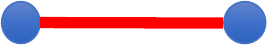}}}}

\usepackage{microtype}
\usepackage{graphicx}
\usepackage{booktabs} 

\usepackage{hyperref}

\usepackage[accepted]{icml2022}

\begin{document}

\twocolumn[
\icmltitle{FMP: Toward Fair Graph Message Passing against Topology Bias}



\icmlsetsymbol{equal}{*}

\begin{icmlauthorlist}
\icmlauthor{Zhimeng Jiang}{tamu}
\icmlauthor{Xiaotian Han}{tamu}
\icmlauthor{Chao Fan}{tamu}
\icmlauthor{Zirui Liu}{rice}
\icmlauthor{Na Zou}{tamu}
\icmlauthor{Ali Mostafavi}{tamu}
\icmlauthor{Xia Hu}{rice}
\end{icmlauthorlist}

\icmlaffiliation{tamu}{Texas A\&M University}
\icmlaffiliation{rice}{Rice University}

\icmlcorrespondingauthor{Zhimeng Jiang}{zhimengj@tamu.edu}
\icmlcorrespondingauthor{Xia Hu}{xia.hu@rice.edu}

\icmlkeywords{Machine Learning, ICML}

\vskip 0.3in
]



\printAffiliationsAndNotice{}  

\begin{abstract}
Despite recent advances in achieving fair representations and predictions through regularization, adversarial debiasing, and contrastive learning in graph neural networks (GNNs), the working mechanism (i.e., message passing) behind GNNs inducing unfairness issue remains unknown. In this work, we theoretically and experimentally demonstrate that representative aggregation in message passing schemes accumulates bias in node representation due to topology bias induced by graph topology. Thus, a \textsf{F}air \textsf{M}essage \textsf{P}assing (FMP) scheme is proposed to aggregate useful information from neighbors but minimize the effect of topology bias in a unified framework considering graph smoothness and fairness objectives. The proposed FMP is effective, transparent, and
compatible with back-propagation training. An acceleration approach on gradient calculation is also adopted to improve algorithm efficiency. Experiments on node classification tasks demonstrate that the proposed FMP outperforms the state-of-the-art baselines in effectively and efficiently mitigating bias on three real-world datasets. 
\end{abstract}


\section{Introduction}
\label{sect:intro}
Graph neural networks (GNNs) \cite{kipf2017semi,velivckovic2018graph,wu2019simplifying} are widely adopted in various domains, such as social media mining \cite{hamilton2017inductive}, knowledge graph \cite{hamaguchi2017knowledge} and recommender system \cite{ying2018graph}, due to remarkable performance in learning representations. Graph learning, a topic with growing popularity, aims to learn node representation containing both topological and attribute information in a given graph. Despite the outstanding performance in various tasks, GNNs still inherit or even amplify societal bias from input graph data \cite{dai2021say}.
The biased node representation largely limits the application of GNNs in many high-stake tasks, such as job hunting \cite{mehrabi2021survey} and crime ratio prediction \cite{suresh2019framework}. Hence, bias mitigation that facilitates the research on fair GNNs is in an urgent need.

In many graphs, nodes with the same sensitive attribute (e.g., ages) are more likely to get connected. 
For example, young people mainly make friends with people at similar ages \cite{dong2016young}.
We call this phenomenon ``topology bias''.
Since, for each node, the node representations is learned by aggregating the representation of its neighbors in GNNs, nodes with the same sensitive attribute will be more similar after the aggregation.
Specifically, we visualize the topology bias for three real-world datasets (Pokec-n, Pokec-z, and NBA) in Figure~\ref{fig:visual}, where different edge types are highlighted with different colors for top-$3$ largest connected components in original graphs.
Such topology bias leads to more similar node representation for those nodes with the same sensitive attribute, which is a major source for the graph representation bias.

Many existing works achieving fair node representation either rely on regularization, adversarial debiasing, or contrastive learning. These methods, however, adopt sensitive attribute information in training loss refinement, ignoring the uniqueness part (i.e., message passing) of GNNs on exploring fairness. The effect of message passing on improving fairness by eliminating topology bias is underexplored in prior works. A natural question is raised: \textbf{Can message passing in GNNs aggregate useful information from neighbors and fundamentally mitigate representation bias inherited from topology bias?}

In this work, we provide a positive answer via designing an effective, efficient, and transparent scheme for GNNs, called fair message passing (FMP). 
First, we theoretically prove that the aggregation in message passing inevitably accumulates representation bias when large topology bias exists. Second, motivated by our theoretical analysis, 
we plug the sensitive attribute information into the message passing scheme.
Specifically, we formulate an optimization problem that integrates fairness and prediction performance objectives.
Then we derive FMP to solve the problem via Fenchel conjugate and gradient descent to generate fair-and-predictive representation. Further, we demonstrate the superiority of FMP by examining its effectiveness and efficiency, where we adopt the property of softmax function to accelerate the gradient calculation over primal variables.
In short, the contributions can be summarized as follows:
\begin{itemize}[leftmargin=0.2cm, itemindent=.0cm, itemsep=0.0cm, topsep=0.0cm]
    \item To the best of our knowledge, it is the first paper to theoretically investigate how GCN-like message passing scheme amplifies representation bias according to topology bias. 
    \item Integrated in a unified optimization framework, we develop the FMP scheme to simultaneously guarantee graph smoothness and enhance fairness, which lays a foundation for intrinsic methods on exploring fair GNNs. An acceleration method is proposed to reduce gradient computational complexity with theoretical and empirical validation. 
    \item  The effectiveness and efficiency of FMP are experimentally evaluated on three real-world datasets. The results show that compared to the state-of-the-art, our FMP exhibits a superior trade-off between prediction performance and fairness with negligibly computation overhead. 
\end{itemize}

\begin{figure}[t]
\centering
\includegraphics[width=0.85\linewidth]{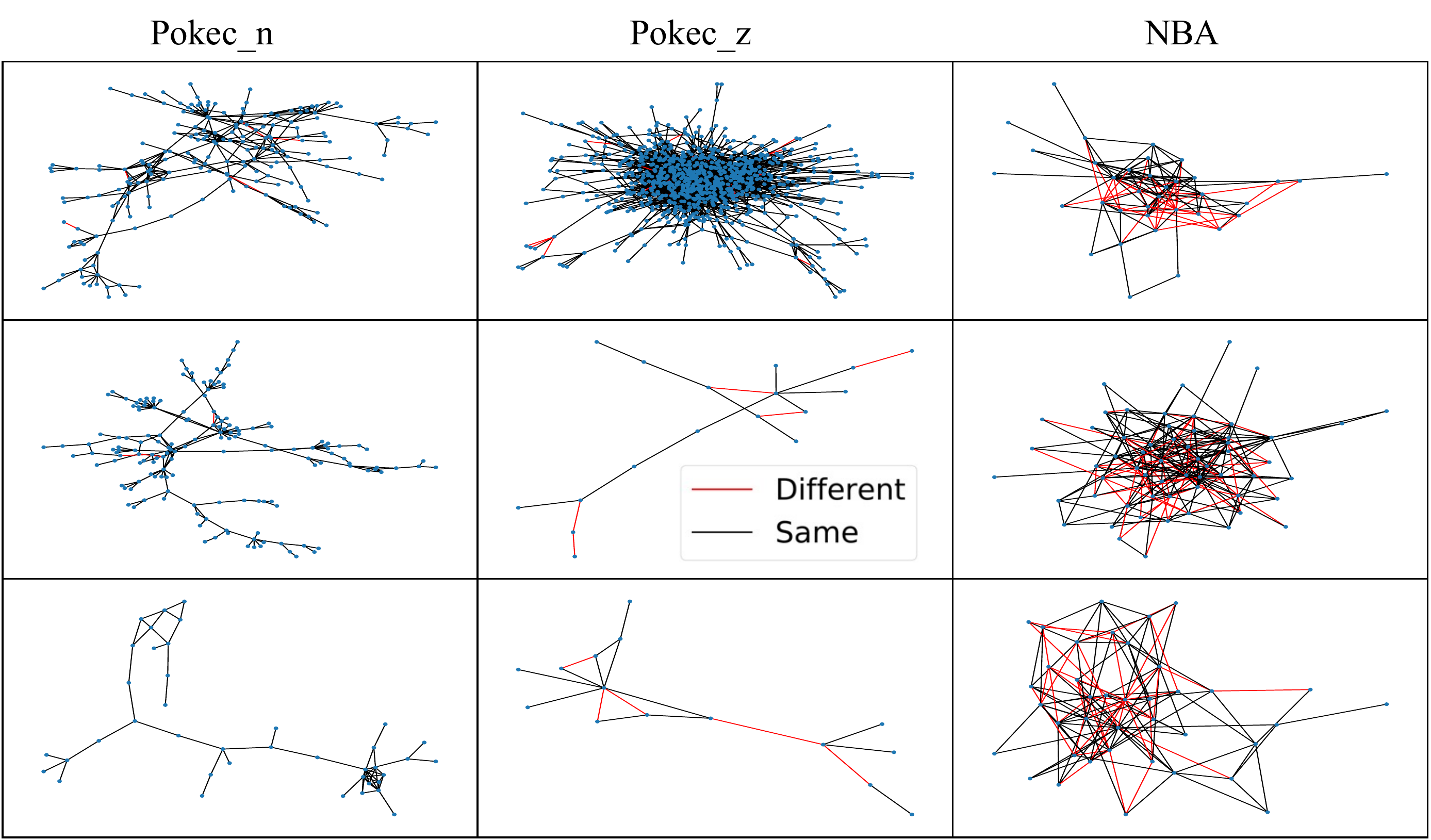}

\caption{Visualization of topology bias in three real-world datasets, where black edges $\blink$ and red edges $\rlink$ represent the edge with the same or different sensitive attributes for the connected node pair, respectively. We visualize the largest three connected components for each dataset. It is obvious that the sensitive homophily coefficients (the ratio of homo edges) are high in practice, i.e., $95.30\%$, $95.06\%$ and $72.37\%$ for Pokec-n, Pokec-z and NBA dataset, respectively.}
\vspace{-15pt}
\label{fig:visual}
\end{figure}

\section{Preliminaries}
We adopt bold upper-case letters to denote matrix such as $\mathbf{X}$, bold lower-case letters such as $\mathbf{x}$ to denote vectors, calligraphic font such as $\mathcal{X}$ to denote set. Given a matrix $\mathbf{X}\in\mathbb{R}^{n\times d}$, the $i$-th row and $j$-th column are denoted as $\mathbf{X}_i$ and $\mathbf{X}_{\cdot,j}$, and the element in $i$-th row and $j$-th column is 
$\mathbf{X}_{i,j}$. We use the Frobenius norm, $l_1$ norm of matrix $\mathbf{X}$ as $||\mathbf{X}||_F=\sqrt{\sum_{i,j}\mathbf{X}_{i,j}^2}$ and $||\mathbf{X}||_1=\sum_{ij}|\mathbf{X}_{ij}|$, respectively. Given two matrices $\mathbf{X}, \mathbf{Y}\in\mathbb{R}^{n \times d}$, the inner product is defined as $\langle\mathbf{X}, \mathbf{Y}\rangle=tr(\mathbf{X}^{\top}\mathbf{Y})$, where $tr(\cdot)$ is the trace of a square matrix. $SF(\mathbf{X})$ represents softmax function with a default normalized column dimension.

Let $\mathcal{G}=\{\mathcal{V}, \mathcal{E}\}$ be a graph with the node set $\mathcal{V}=\{v_1, \cdots, v_n\}$ and the undirected edge set $\mathcal{E}=\{e_1, \cdots, e_m\}$, where $n, m$ represent the number of node and edge, respectively. The graph structure $\mathcal{G}$ can be represented as an adjacent matrix $\mathbf{A}\in\mathbb{R}^{n\times n}$, where $\mathbf{A}_{ij}=1$ if existing edge between node $v_i$ and node $v_j$. $\mathcal{N}(i)$ denotes the neighbors
of node $v_i$ and $\tilde{\mathcal{N}}(i)=\mathcal{N}(i)\cup \{v_i\}$ denotes the self-inclusive neighbors.
Suppose that each node is associated with a $d$-dimensional feature vector and a (binary) sensitive attribute, the feature for all nodes and sensitive attribute are denoted as $\mathbf{X}_{ori}=\mathbb{R}^{n\times d}$ and $\mathbf{s}\in \{-1, 1\}^{n}$. Define the sensitive attribute incident vector as $\Delta_{\mathbf{s}}= \frac{\mathbbm{1}_{>0}(\mathbf{s})}{||\mathbbm{1}_{>0}(\mathbf{s})||_1} - \frac{\mathbbm{1}_{>0}(-\mathbf{s})}{||\mathbbm{1}_{>0}(-\mathbf{s})||_1}$ to normalize each sensitive attribute group, where $\mathbbm{1}_{>0}(\mathbf{s})$ is an element-wise indicator function.

\subsection{GNNs as Graph Signal Denoising}
A GNN model is usually composed of several stacking GNN layers. Given a graph $\mathcal{G}$ with $N$ nodes, a GNN layer typically contains feature transformation $\mathbf{X}_{trans}=f_{trans}(\mathbf{X}_{ori})$ and aggregation $\mathbf{X}_{agg}=f_{agg}(\mathbf{X}_{trans}|\mathcal{G})$, where $\mathbf{X}_{ori}\in\mathbb{R}^{n\times d_{in}}$, $\mathbf{X}_{trans}, \mathbf{X}_{agg}\in\mathbb{R}^{n\times d_{out}}$ represent the input and output features. The feature transformation operation transforms the node feature dimension via neural networks, and \emph{feature aggregation}, updates node features based on neighbors' features and graph topology.

Recent works \cite{ma2021unified, zhu2021interpreting} have established the connections between many feature aggregation operations in representative GNNs and a graph signal denoising problem with Laplacian regularization. Here, we only introduce GCN/SGC as an examples to show the connection from a perspective of graph signal denoising. The more discussions are elaborated in  \underline{Appendix~\ref{app:denoise}}. Feature aggregation in Graph Convolutional Network (GCN) or Simplifying Graph Convolutional Network (SGC) is given by $\mathbf{X}_{agg}=\tilde{\mathbf{A}}\mathbf{X}_{trans}$, where $\tilde{\mathbf{A}}=\tilde{\mathbf{D}}^{-\frac{1}{2}}\hat{\mathbf{A}}\tilde{\mathbf{D}}^{-\frac{1}{2}}$ is a normalized self-loop adjacency matrix $\hat{\mathbf{A}}=\mathbf{A}+\mathbf{I}$, and $\tilde{\mathbf{D}}$ is degree matrix of $\tilde{\mathbf{A}}$. Recent works \cite{ma2021unified, zhu2021interpreting} provably demonstrate that such feature aggregation is equivalent to one-step gradient descent to minimize $tr(\mathbf{F}^{\top}\big(\mathbf{I}-\tilde{\mathbf{A}})\mathbf{F}\big)$ with initialization $\mathbf{F}=\mathbf{X}_{trans}$.

\subsection{Label and Sensitive Homophily in Graphs}
The behaviors of graph neural networks have been investigated in the context of label homophily for connected node pairs in graphs
\cite{ma2021homophily}. Label homophily in graphs is typically defined to characterize the similarity of connected node labels in graphs. Here, similar node pair means that the connected nodes share the same label. From the perspective of fairness, we also define the sensitive homophily coefficient to represent the sensitive attribute similarity among connected node pairs. Informally, the coefficient for label homophily and sensitive homophily are defined as the fraction of the edges connecting the nodes of the same class label and sensitive attributes in a graph \cite{zhu2020beyond, ma2021homophily}. We also provide the formal definition as follows:
\begin{definition}[Label and Sensitive Homophily Coefficient]
Given a graph $\mathcal{G} = \{\mathcal{V}, \mathcal{E}\}$ with node label vector $\mathbf{y}$, node sensitive attribute vector $\mathbf{s}$, the label and sensitive attribute homophily
coefficients are defined as the fraction of edges that connect nodes with the same labels or sensitive attributes
$\epsilon_{label}(\mathcal{G}, \mathbf{y})= \frac{1}{|\mathcal{E}|}\sum_{(i,j)\in \mathcal{E}} \mathbbm{1}(\mathbf{y}_i = \mathbf{y}_j)$,  and
$\epsilon_{sens}(\mathcal{G}, \mathbf{s})=
\frac{1}{|\mathcal{E}|}\sum_{(i,j)\in \mathcal{E}} \mathbbm{1}(\mathbf{s}_i = \mathbf{s}_j)$,
where $|\mathcal{E}|$ is the number of edges and $\mathbbm{1}(\cdot)$ is the indicator function. 
\end{definition}

Recent works \cite{ma2021homophily,chien2021adaptive,zhu2020beyond} aim to understand the relation between the message passing in GNNs and label homophily from the interactions between GNNs (model) and graph topology (data). For graph data with a high label homophily coefficient, existing works \cite{ma2021homophily, tang2020investigating} have demonstrated, either provably or empirically, that the nodes with higher node degree obtain more prediction benefits in GCN, compared to the benefits that peripheral nodes obtain. As for graph data with a low label homophily coefficient, GNNs does not necessarily lead to better prediction performance compared with MLP since since the node features of neighbors with different labels contaminates the node features during feature aggregation. However, although work \cite{dai2021say} empirically points out that graph data with large sensitive homophily coefficient may enhance bias inGNNs, 
the fundamental understanding for message passing and sensitive homophily coefficient is still missing. We provide the theoretical analysis in Section~\ref{sect:enhance_bias}.

\section{Graph Topology Enhances Bias in GNNs} \label{sect:enhance_bias}
For each node, GNNs aggregate its neighbors' features to learn its representation. In real-world datasets, we observe a high sensitive homophily coefficient (which is even higher than the label homophily coefficient). However, existing message passing schemes tend to aggregate the node features from their neighbors with the same sensitive attributes. 
Thus, the common belief is that the message passing renders node representations with the same sensitive attribute more similar. 
However, this common belief is heuristic and the quantifiable relationship between the topology bias and representation bias is still missing. In this section, (1) we rectify such common belief and quantitatively reveal that \textbf{only sufficient high sensitive homophily coefficient} would lead to bias enhancement; (2) we analyze the \textbf{influence} of other graphs statistical information, such as \textbf{the number of nodes $n$, the edge density $\rho_{d}$, and sensitive group ratio $c$}, in term of bias enhancement.

Throughout our analysis, we mainly focus on $4$ characteristics of graph topology: \textbf{the number of nodes $n$, the edge density $\rho_{d}$, sensitive homophily coefficient $\epsilon_{sens}$, and sensitive group ratio $c$}. Specifically, we consider the synthetic random graph generation as follows: 
\begin{definition}[$(n, \rho_{d}, \epsilon_{sens}, c)$-graph]
The synthetic random graph $\mathcal{G}$ sampled from $(n, \rho_{d}, \epsilon_{sens}, c)$-graph satisfies the following properties: 1) the graph node number is $n$; 2) the adjacency matrix $\mathbf{A}$ satisfies $\mathbf{A}_{ij}\in\{0,1\}$ and $\mathbb{E}_{ij}[\mathbb{P}(\mathbf{A}_{ij}=1)]=\rho_{d}$; 3) given connected node pair $\mathbf{A}_{ij}=1$, the probability of connected nodes with the same sensitive attribute satisfies $\mathbb{P}(\mathbf{s}_i=\mathbf{s}_j|\mathbf{A}_{ij}=1)=\epsilon_{sens}$; 4) the binary sensitive attribute $\mathbf{s}_i\in\{-1,1\}$ satisfies $\mathbb{E}_i[\mathbb{P}(\mathbf{s}_i=1)]=c$; 5) independent edge generation.
\end{definition}
We assume that node attributes in synthetic graph follow Gaussian Mixture Model $GMM(c, \mathbf{\mu}_1, \mathbf{\Sigma}_1, \mathbf{\mu}_2, \mathbf{\Sigma}_2)$. For node with sensitive attribute $\mathbf{s_i}=-1$ ($\mathbf{s_i}=1$), the node attributes $\mathbf{X}_i$ follows Gaussian distribution $P_1 = \mathcal{N}(\mathbf{\mu}_1, \mathbf{\Sigma}_1)$ ($P_2 = \mathcal{N}(\mathbf{\mu}_2, \mathbf{\Sigma}_2)$), where the node attributes with the same sensitive attribute are independent and identically distributed, and $\mathbf{\mu}_{i}, \mathbf{\Sigma}_i$ ($i=1,2$) represent the mean vector and covariance matrix. 

To measure the node representations bias, we adopt the mutual information between sensitive attribute and node attributes $I(\mathbf{s}, \mathbf{X})$. Note that the exact mutual information $I(\mathbf{s}, \mathbf{X})$ is intractable to estimate, an upper bound on the exact mutual information is developed as a surrogate metric in the following Theorem~\ref{theo:surro}:
\begin{theorem}\label{theo:surro}
Suppose the synthetic graph node attribute $\mathbf{X}$ is generated based on Gaussian Mixture Model $GMM(c, \mathbf{\mu}_1, \mathbf{\Sigma}_1, \mathbf{\mu}_2, \mathbf{\Sigma}_2)$, i.e., the probability density function of node attributes  for the nodes of different sensitive attribute $\mathbf{s}=\{-1, 1\}$ follows $f_{\mathbf{X}}(\mathbf{X}_i=\mathbf{x}|\mathbf{s}_i=-1)\sim \mathcal{N}(\mathbf{\mu}_1, \mathbf{\Sigma}_1)\dff P_1$ and $f_{\mathbf{X}}(\mathbf{X}_i=\mathbf{x}|\mathbf{s}_i=1)\sim \mathcal{N}(\mathbf{\mu}_2, \mathbf{\Sigma}_2)\dff P_2$, and the sensitive attribute ratio satisfies $\mathbb{E}_i[\mathbb{P}(\mathbf{s}_i=1)]=c$, then the mutual information between sensitive attribute and node attributes $I(\mathbf{s}, \mathbf{X})$ satisfies 
\be 
I(\mathbf{s}, \mathbf{X})\leq -(1-c)\ln\Big[(1-c) +c\exp\big(-D_{KL}(P_1||P_2)\big)\Big] \nonumber\\
-c\ln\Big[c +(1-c)\exp\big(-D_{KL}(P_2||P_1)\big)\Big]
\dff Bias(\mathbf{s}, \mathbf{X}). \nonumber
\ee 
\end{theorem}
Based on Theorem~\ref{theo:surro}, we can observe that lower distribution distance $D_{KL}(P_1||P_2)$ or $D_{KL}(P_2||P_1)$ is beneficial for reducing $Bias(\mathbf{s}, \mathbf{X})$ and $I(\mathbf{s}, \mathbf{X})$ since the sensitive attribute is less distinguishable based on node representations. 

We focus on the role of message passing in terms of fairness. Suppose the graph adjacency matrix $\mathbf{A}$ is sampled for $(n, \rho_{d}, \epsilon_{sens}, c)$-graph and we adopt the GCN-like message passing $\tilde{\mathbf{X}}=\tilde{\mathbf{A}}\mathbf{X}$, where $\tilde{\mathbf{A}}$ is normalized adjacency matrix with self-loop. We define the bias difference for such message passing as $\Delta Bias=Bias(\mathbf{s}, \tilde{\mathbf{X}})-Bias(\mathbf{s}, \mathbf{X})$ to measure the role of graph topology. Subsequently, we provide a sufficient condition to specify the case that graph topology enhances bias in Theorem~\ref{theo:enhance}.
\begin{theorem}\label{theo:enhance}
Suppose the synthetic graph node attribute $\mathbf{X}$ is generated based on Gaussian Mixture Model $GMM(c, \mathbf{\mu}_1, \mathbf{\Sigma}, \mathbf{\mu}_2, \mathbf{\Sigma})$, and the graph adjacency matrix $\mathbf{A}$ is generated from $(n, \rho_{d}, \epsilon_{sens}, c)$-graph. If adopting GCN-like message passing $\tilde{\mathbf{X}}=\tilde{\mathbf{A}}\mathbf{X}$, bias will be enhanced, i.e., $\Delta Bias>0$ if the bias-enhance condition holds: $(\nu_1-\nu_2)^2\min\{\zeta_1, \zeta_2\} > 1$,
where $\nu_1-\nu_2<1$ represents the reduction coefficient of the distance between the mean node attributes of the two sensitive attribute group, $\zeta_1, \zeta_2$ mean the connection degree of two sensitive group; the mathematical formulation is given by
\be 
\nu_1 = \frac{(n_1-1)p_{conn}+1}{\zeta_0}, \quad 
\nu_2 = \frac{(n_1-1)q_{conn}}{\zeta_1}\nonumber\\
\zeta_1=n_{-1}q_{conn}+(n_1-1)p_{conn}+1, \nonumber\\
\zeta_2 = n_{-1}p_{conn}+(n_1-1)q_{conn}+1
\ee 
where the node number with the same sensitive attribute $n_{-1}=n(1-c)$, $n_1=nc$, intra-connect probability $p_{conn}=\mathbb{E}_{ij}[\mathbb{P}(\mathbf{A}_{ij}|\mathbf{s}_i=\mathbf{s}_j)]$, and inter-connect probability $q_{conn}=\mathbb{E}_{ij}[\mathbb{P}(\mathbf{A}_{ij}|\mathbf{s}_i\mathbf{s}_j=-1)]$.
\end{theorem}
Based on Theorem~\ref{theo:enhance}, we have the following theoretical observations (See more elaboration in \underline{Appendix~\ref{app:discussions})}:
\begin{itemize}[leftmargin=0.2cm, itemindent=.0cm, itemsep=0.0cm, topsep=0.0cm]
    \item \textbf{Node representation bias is enhanced by message passing for sufficient large sensitive homophily $\epsilon_{sens}$.} When $\epsilon_{sens}$ approaches $1$, the inter-connect probability is $0$. Hence the mean reduction coefficient is $\nu_1-\nu_2=1$ and connection degree $\min\{\zeta_1, \zeta_2\} > 1$.
    \item \textbf{The bias enhancement implicitly depends on node representation geometric differentiation, including the distance between the mean node representation within the same sensitive attribute and the scale covariance matrix.} Theorem~\ref{theo:surro} implies that low mean representation distance and concentrated representation (low covariance matrix) leads to fair representation. However,
    GCN-like message passing renders the mean node representation distance reduction $\nu_1-\nu_2$ and concentrated for each sensitive attribute group, which is an ``adversarial" effect for fairness and the mean distance and covariance reduction is controlled by sensitive homophily coefficient. 
    \item \textbf{The bias is enlarged as node number $n$ being increased.} For large node number $n$, the mean distance almost keeps constant since $\nu_1\approx\frac{c p_{conn}}{(1-c)q_{conn}+c p_{conn}}$, $\nu_2\approx\frac{c q_{conn}}{(1-c)p_{conn}+c q_{conn}}$, and $\zeta_1, \zeta_2$ are almost proportional to node number $n$. Therefore, the bias-enhancement condition can be more easily satisfied in large graphs.
    \item \textbf{The bias is enlarged as graph connection density $\rho$ being increased.} Based on \underline{Appendix~\ref{app:enhance}}, inter-connect and intra-connect probability is proportional to graph connection density $\rho_d$, and the mean distance almostly keeps constant and $\zeta_1, \zeta_2$ are almost proportional to $\rho_d$. Therefore, the bias-enhancement condition can be more easily satisfied in dense graphs.
    \item \textbf{When the sensitive attribute is more balanced (i.e.,) the bias will be enlarged.} The difference of intra-connected probability $p_{conn}$ and inter-connected probability $q_{conn}$ is increased if the sensitive attribute is highly balanced. Therefore, the aggregated node representation will adopt more neighbor's information with the same sensitive attribute during message passing and amplify representation bias.
\end{itemize}

\section{Fair Message Passing}
Section~\ref{sect:enhance_bias} demonstrates that GCN-like message passing enhance node representation bias for graph data with large topology bias. 
In this section, we propose a new message passing scheme to aggregate useful information from neighbors while debiasing representation bias. Specifically, we formulate fair message passing as an optimization problem to pursue \emph{smoothness} and \emph{fair} node representation simultaneously. Together with an effective and efficient optimization algorithm, we derive the closed-form fair message passing. Finally, the proposed FMP is shown to be integrated in fair GNNs at three stages (i.e., transformation, aggregation, and debiasing). 

\subsection{The Optimization Framework}
To achieve graph smoothness prior and fairness in the same process, a reasonable message passing should be a good solution for the following optimization problem:
\be 
\min\limits_{\mathbf{F}} \underbrace{\frac{\lambda_{s}}{2}tr(\mathbf{F}^{T}\tilde{\mathbf{L}}\mathbf{F}) + \frac{1}{2} ||\mathbf{F}-\mathbf{X}_{trans}||^2_{F}}_{h_s(\mathbf{F})} + \underbrace{\lambda_{f}||\mathbf{\Delta}_s SF(\mathbf{F})||_{1}}_{h_f\big(\mathbf{\Delta}_s SF(\mathbf{F})\big)}. \nonumber
\ee  
where $\mathbf{X}_{trans}\in \mathbf{R}^{n \times d_{out}}$ is the transformed $d_{out}$-dimensional node features and $\mathbf{F}\in \mathbf{R}^{n \times d_{out}}$ is the aggregated node features of the same matrix size. The first two terms preserve the similarity of connected node representation and thus enforces graph smoothness. The last term enforces fair node representation so that the average predicted probability between groups of different sensitive attributes can remain constant. The regularization coefficients $\lambda_s$ and $\lambda_f$ adaptively control the trade-off between graph smoothness and fairness. 

\paragraph{Smoothness objective.} The adjacent matrix in existing graph message passing schemes is normalized for improving numerical stability and achieving superior performance. Similarly, the graph smoothness term requires normalized Laplacian matrix, i.e., $\tilde{\mathbf{L}}=\mathbf{I}-\tilde{\mathbf{A}}$, $\tilde{\mathbf{A}}=\hat{\mathbf{D}}^{-\frac{1}{2}}\hat{\mathbf{A}}\hat{\mathbf{D}}^{-\frac{1}{2}}$, and $\hat{\mathbf{A}}=\mathbf{A}+\mathbf{I}$. From edge-centric view, smoothness objective enforces connected node representation to be similar since $tr(\mathbf{F}^{T}\tilde{\mathbf{L}}\mathbf{F})=\sum_{(v_i, v_j)\in\mathcal{E}}||\frac{\mathbf{F}_i}{\sqrt{d_i+1}}-\frac{\mathbf{F}_j}{\sqrt{d_j+1}}||^2_F$, where $d_i=\sum_{k}A_{ik}$ represents the degree of node $v_i$.

\paragraph{Fairness objective.} The fairness objective measure the bias for node representation after aggregation. Recall sensitive attribute incident vector $\Delta_{\mathbf{s}}$ indicates the sensitive attribute group and group size via the sign and absolute value summation. Recall that the sensitive attribute incident vector as $\Delta_{\mathbf{s}}= \frac{\mathbbm{1}_{>0}(\mathbf{s})}{||\mathbbm{1}_{>0}(\mathbf{s})||_1} - \frac{\mathbbm{1}_{>0}(-\mathbf{s})}{||\mathbbm{1}_{>0}(-\mathbf{s})||_1}$ and $SF(\mathbf{F})$ represents the predicted probability for node classification task, where $SF(\mathbf{F})_{ij}=\hat{P}(y_i=j|\mathbf{X})$. Furthermore, we can show that our fairness objective is actually equivalent to demographic parity, i.e., $\Big(\Delta_s SF(\mathbf{F})\big)\Big)_j=\hat{P}(y_i=j|\mathbf{s}_i=1, \mathbf{X}) - \hat{P}(y_i=j|\mathbf{s}_i=-1, \mathbf{X})$. Please see proof in \underline{Appendix~\ref{app:fairnessobj}}.
In other words, our fairness objective, $l_1$ norm of $\Delta_s SF(\mathbf{F})$ characterizes the predicted probability difference between two groups with different sensitive attribute. Therefore, our proposed optimization framework can pursue graph smoothness and fairness simultaneously.

\subsection{Algorithm for Fair Message Passing}
For smoothness objective, many existing popular message passing scheme can be derived based on gradient descent with appropriate step size choice \cite{ma2021unified,zhu2021interpreting}. However, directly computing the gradient of the fairness term makes the closed-form gradient complicated since the gradient of $l_1$ norm involves the sign of elements in the vector. To solve this optimization problem in a more effective and efficient manner, Fenchel conjugate \cite{rockafellar2015convex} is introduced to transform the original problem as bi-level optimization problem (Please see more details in \underline{Appendix~\ref{app:fmp}}). For the general convex function $h(\cdot)$, its conjugate function is defined as 
$h^{*}(\mathbf{U})\dff \sup\limits_{\mathbf{X}}\langle \mathbf{U}, \mathbf{X}\rangle -h(\mathbf{X})$.
Based on Fenchel conjugate, the fairness objective can be transformed as variational representation $h_f(\mathbf{p})=\sup\limits_{\mathbf{u}}\langle\mathbf{p},\mathbf{u} \rangle - h_f^{*}(\mathbf{u})$, where $\mathbf{p}=\mathbf{\Delta}_s SF(\mathbf{F})\in\mathbb{R}^{1\times d_{out}}$ is a predicted probability vector for classification. Furthermore, the original optimization problem is equivalent to 
\be \label{eq:minmax}
\min\limits_{\mathbf{F}}\max\limits_{\mathbf{u}} h_s(\mathbf{F}) + \langle\mathbf{p},\mathbf{u} \rangle - h_f^{*}(\mathbf{u})
\ee
where $\mathbf{u}\in\mathbb{R}^{1\times d_{out}}$ and $h_f^{*}(\cdot)$ is the conjugate function of fairness objective $h_f(\cdot)$.

Motivated by Proximal Alternating Predictor-Corrector (PAPC) \cite{loris2011generalization,chen2013primal}, we develop an efficient algorithm  to solve bi-level optimization problem (\ref{eq:minmax}) with low computation complexity. Specifically, we adopt predict-then-correct algorithm to solve bi-level optimization in a close-form manner. We summarize the proposed FMP as two phases, including propagation with skip connection (Step \textbf{\ding{182}}) and bias mitigation (Steps \textbf{\ding{183}}-\textbf{\ding{186}}). For bias mitigation, Step \textbf{\ding{183}} updates the aggregated node features for fairness objective; Steps \textbf{\ding{184}} and \textbf{\ding{185}} aim to learn and ``reshape" perturbation vector in probability space, respectively. Step \textbf{\ding{186}} explicitly mitigate the bias of node features based on gradient descent on primal variable. The mathematical formulation is given as follows:
\be 
\left\{
\begin{array}{ll}
\mathbf{X}_{agg}^{k+1}=\gamma \mathbf{X}_{trans} + (1-\gamma)\tilde{\mathbf{A}}\mathbf{X}^{k}, & \text{Step \textbf{\ding{182}}}\\
\bar{\mathbf{F}}^{k+1}=\mathbf{X}_{agg}^{k+1}-\gamma \frac{\partial \langle \mathbf{p}, \mathbf{u}^{k}\rangle}{\partial \mathbf{F}}\Big|_{\mathbf{F}^{k}}, & \text{Step \textbf{\ding{183}}}\\
\bar{\mathbf{u}}^{k+1}=\mathbf{u}^{k}+\beta \mathbf{\Delta_{s}} SF(\bar{\mathbf{F}}^{k+1}), & \text{Step \textbf{\ding{184}}}\\
\mathbf{u}^{k+1}=\min\Big(|\bar{\mathbf{u}}^{k+1}|, \lambda_{f} \Big)\cdot sign(\bar{\mathbf{u}}^{k+1}), & \text{Step \textbf{\ding{185}}}\\
\mathbf{F}^{k+1}=\mathbf{X}_{agg}^{k+1}-\gamma \frac{\partial \langle \mathbf{p}, \mathbf{u}^{k+1}\rangle}{\partial \mathbf{F}}\Big|_{\mathbf{F}^{k}}. & \text{Step \textbf{\ding{186}}}
\end{array}
\right. \nonumber
\ee 
where $\gamma=\frac{1}{1+\lambda_f}$, $\beta=\frac{1}{2\gamma}$, $\mathbf{X}_{agg}^{k+1}$ represents the node features with normal aggregation and skip connection with the transformed input $\mathbf{X}_{trans}$.

\begin{figure}[t]
\centering
\includegraphics[width=0.99\linewidth]{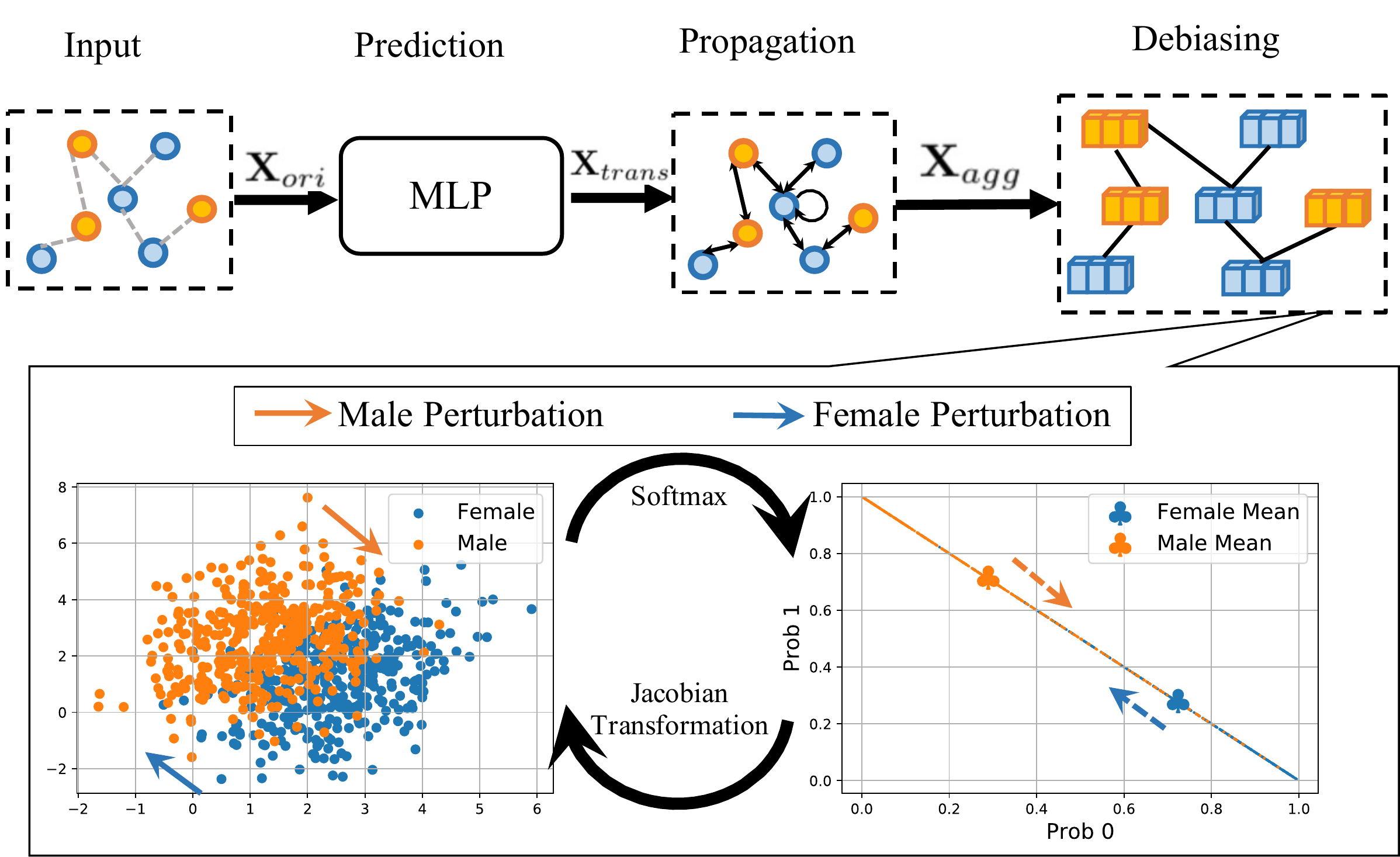}

\caption{The model pipeline consists of three steps: MLP (feature transformation), propagation with skip connection and debiasing via low-rank perturbation in probability space. }
\vspace{-15pt}
\label{fig:illu}
\end{figure}

\begin{figure*}[t]
\centering
\includegraphics[width=0.99\linewidth]{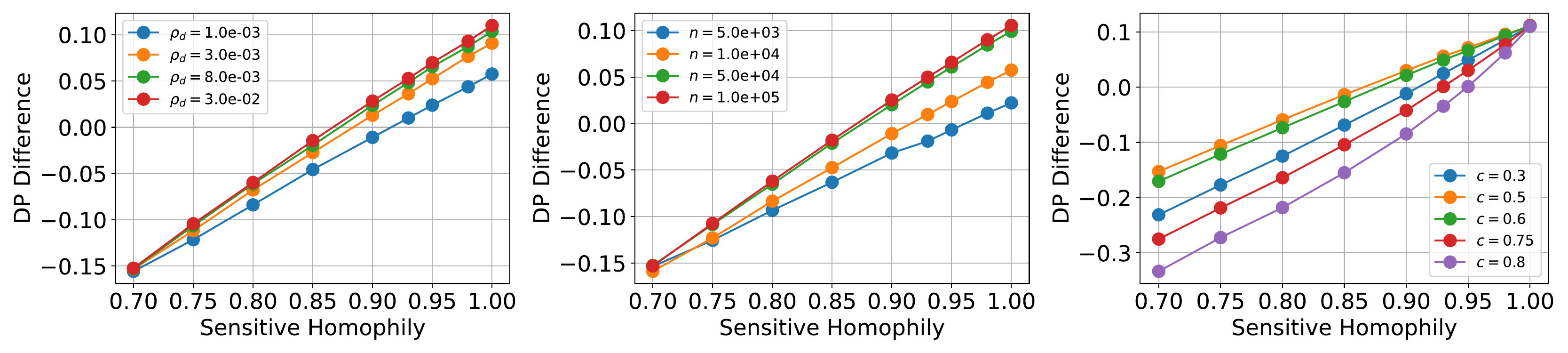}
\caption{The difference of demographic parity for message passing. \textbf{Left:} DP difference for different graph connection density $\rho_d$ with senstive attribute ratio $c=0.5$ and number of nodes $n=10^{4}$; \textbf{Middle:} DP difference for different number of nodes $n$ with senstive attribute ratio $c=0.5$ and graph connection density $\rho_d=10^{-3}$; \textbf{Right:} DP difference for different senstive attribute ratio $c$ with graph connection density $\rho_d=10^{-3}$ and number of nodes $n=10^{4}$;
}
\vspace{-5pt}
\label{fig:dp_diff}
\end{figure*}

\paragraph{Gradient Computation.} The softmax property is also adopted to accelerate the gradient computation. Note that $\mathbf{p}=\mathbf{\Delta}_s SF(\mathbf{F})$ and $SF(\cdot)$ represents softmax over column dimension, directly computing the gradient $\frac{\partial \langle \mathbf{p}, \mathbf{u}\rangle}{\partial \mathbf{F}}$ based on chain rule involves the three-dimensional tensor $\frac{\partial \mathbf{p}}{\partial \mathbf{F}}$ with gigantic computation complexity. Instead, we simplify the gradient computation based on the property of softmax function in the following theorem. 

\begin{theorem}[Gradient Computation]\label{theo:grad_comp}
The gradient over primal variable $\frac{\partial \langle \mathbf{p}, \mathbf{u}\rangle}{\partial \mathbf{F}}$ satisfies 
\be 
\frac{\partial \langle \mathbf{p}, \mathbf{u}\rangle}{\partial \mathbf{F}}= \mathbf{U}_s\odot SF(\mathbf{F})-\text{Sum}_{1}(\mathbf{U}_s\odot SF(\mathbf{F}))SF(\mathbf{F}).
\ee 
where $\mathbf{U}_s\dff \Delta_s^{\top}\mathbf{u}$, $\odot$ represents element-wise product and $\text{Sum}_{1}(\cdot)$ represents the summation over column dimension with preserved matrix shape. 
\end{theorem}

\subsection{FMP Analysis}
In this section, we analyze the \emph{transparency} and \emph{efficiency} and \emph{effectiveness} of proposed FMP scheme. Specifically, we intreprete FMP as conventional aggregation plus bias mitigation two phrases, and the computation complexity of FMP is lower than backward gradient calculation. 
\paragraph{Transparency} Note that the gradient of fairness objective over node features $\mathbf{F}$ satisfies $\frac{\partial \langle \mathbf{p}, \mathbf{u}\rangle}{\partial \mathbf{F}}=\frac{\partial \langle \mathbf{p}, \mathbf{u}\rangle}{\partial SF(\mathbf{F})}\frac{\partial SF(\mathbf{F})}{\partial \mathbf{F}}$ and $\frac{\partial \langle \mathbf{p}, \mathbf{u}\rangle}{\partial SF(\mathbf{F})}=\Delta_s^{\top}\mathbf{u}$, such gradient calculation can be interpreted as three steps: Softmax transformation, pertubation in probability space, and debiasing in representation space. Specifcally, we first map the node representation into probability space via softmax transformation. Subsequently, we calculate the gradient of fairness objective in probability space. It is seen that the pertubation $\Delta_s^{\top}\mathbf{u}$ actually poses \emph{low-rank} debiasing in probability space, where the nodes with different sensitive attribute embrace opposite pertubation. In other words, \emph{the dual variable $\mathbf{u}$ represents the pertubation direction in probability space.}
Finally, the pertubation in probability space will be transformed into representation space via Jacobian transformation $\frac{\partial SF(\mathbf{F})}{\partial \mathbf{F}}$.

\paragraph{Efficiency} FMP is an efficient message passing scheme. The computation complexity for the aggregation (sparse matrix multiplications) is $O(md_{out})$, where $m$ is the number of edges in the graph.
For FMP, the extra computation mainly focus on the perturbation calculation, as shown in Theorem~\ref{theo:grad_comp}, with the computation complexity $O(nd_{out})$. The extra computation complexity is negligible in that the number of nodes $n$ is far less than the number of edges $m$ in the real-world graph. Additionally, if directly adopting autograd to calculate the gradient via back propagation, we have to calculate the three-dimensional tensor $\frac{\partial \mathbf{p}}{\partial \mathbf{F}}$ with computation complexity $O(n^2d_{out})$. In other words, 
thanks to the softmax property, we achieve an efficient fair message passing scheme.

\paragraph{Effectiveness} The proposed FMP explicitly achieves graphs smoothness prior and fairness via alternative gradient descent. In other words, the propagation and debiasing forward in a white-box manner and there is not any trainable weight during forwarding phrase. The effectiveness of the proposed FMP is also validated in the experiments on three real-world datasets.

\begin{table*}[h]
\footnotesize
\begin{center}
\caption{Comparative Results with Baselines on Node Classification.}
\label{table:comp_gnns}
\scalebox{0.85}{
    \begin{tabular}{ c|ccc|ccc|ccc} 
    \toprule
     \multirow{2}*{Models} & \multicolumn{3}{c}{Pokec-z} & \multicolumn{3}{c}{Pokec-n} & \multicolumn{3}{c}{NBA} \\
    \cline{2-10}
     & Acc ($\%$) $\uparrow$ & $\Delta_{DP}$ ($\%$) $\downarrow$ & $\Delta_{EO}$ ($\%$) $\downarrow$ & Acc ($\%$) $\uparrow$ & $\Delta_{DP}$ ($\%$) $\downarrow$ & $\Delta_{EO}$ ($\%$) $\downarrow$ & Acc ($\%$) $\uparrow$ & $\Delta_{DP}$ ($\%$) $\downarrow$ & $\Delta_{EO}$ ($\%$) $\downarrow$ \\
    \hline
    MLP &  70.48 $\pm$ 0.77 & 1.61 $\pm$ 1.29 & 2.22 $\pm$ 1.01 & 72.48 $\pm$ 0.26 & 1.53 $\pm$ 0.89 & 3.39 $\pm$ 2.37 & 65.56 $\pm$ 1.62 & 22.37 $\pm$ 1.87 & 18.00 $\pm$ 3.52 \\
    \hline
    GAT &  69.76 $\pm$ 1.30 & 2.39 $\pm$ 0.62 & 2.91 $\pm$ 0.97 & 71.00 $\pm$ 0.48 & 3.71 $\pm$ 2.15 & 7.50 $\pm$ 2.88 & 57.78 $\pm$ 10.65 & 20.12 $\pm$ 16.18 & 13.00 $\pm$ 13.37 \\
    \hline
    GCN & 71.78 $\pm$ 0.37 & 3.25 $\pm$ 2.35 & 2.36 $\pm$ 2.09 & 73.09 $\pm$ 0.28 & 3.48 $\pm$ 0.47 & 5.16 $\pm$ 1.38 & 61.90 $\pm$ 1.00 & 23.70 $\pm$ 2.74 & 17.50 $\pm$ 2.63 \\
    \hline
    SGC & 71.24 $\pm$ 0.46 & 4.81 $\pm$ 0.30 & 4.79 $\pm$ 2.27 & 71.46 $\pm$ 0.41 & 2.22 $\pm$ 0.29 & 3.85 $\pm$ 1.63 & 63.17 $\pm$ 0.63 & 22.56 $\pm$ 3.94 & 14.33$\pm$ 2.16 \\
    \hline
    APPNP & 66.91 $\pm$ 1.46 & 3.90 $\pm$ 0.69 & 5.71 $\pm$ 1.29 & 69.80 $\pm$ 0.89 & 1.98 $\pm$ 1.30 & 4.01 $\pm$ 2.36 & 63.80 $\pm$ 1.19 & 26.51 $\pm$ 3.33 & 20.00 $\pm$ 4.56 \\
    \bottomrule
    FMP & \textbf{70.50} $\pm$ 0.50 & \textbf{0.81} $\pm$ 0.40 & \textbf{1.73} $\pm$ 1.03 & 72.16 $\pm$ 0.33 & \textbf{0.66} $\pm$ 0.40 & \textbf{1.47} $\pm$ 0.87 & \textbf{73.33} $\pm$ 1.85 & \textbf{18.92} $\pm$ 2.28 & \textbf{13.33} $\pm$ 5.89 \\
    \bottomrule
    \end{tabular}
}
\end{center}
\vspace{-10pt}
\end{table*}

\section{Experiments}
In this section, we conduct experiments to validate the effectiveness and efficiency (see \underline{Apppendix~\ref{app:more_exp}} for more details.) of proposed FMP. We firstly validate that graph data with large sensitive homophily enhances bias in GNNs via synthetic experiments. Moreover, for experiments on real-world datasets, we
introduce the experimental settings and then evaluate our proposed FMP compared with several baselines interms of prediction performance and fairness metrics. 

\subsection{Synthetic Experiments} \label{subsect:syn}
We investigate the influence of graph node number $n$, graph connection density $\rho_d$, sensitive homophily $\epsilon_{sens}$, and sensitive attribute ration $c$ for bias enhancement of GCN-like message passing. \textbf{For evaluation metric,}
we adopt the demographic parity (DP) difference during message passing to measure the bias enhancement. For \textbf{node attribute generation}, we first generate node attribute with Gaussian distribution $\mathcal{N}(\mathbf{\mu}_1, \mathbf{\Sigma})$ and $\mathcal{N}(\mathbf{\mu}_2, \mathbf{\Sigma})$ for node with binary sensitive attribute, respectively, where $\mathbf{\mu}_1=[0,1]$, $\mathbf{\mu}_1=[1,0]$ and $\mathbf{\Sigma}=\left[ {\begin{array}{cc}
   1 & 0 \\
   0 & 1 \\
  \end{array} } \right]$. 
\textbf{For adjacency matrix generation}, we randomly generate edges via stochastic block model based on the intra-connection and inter-connection probability.

Figure\ref{fig:dp_diff} shows the DP difference during message passing with respect to sensitive homophily coefficient. We observe that higher sensitive homophily coefficient generally leads to larger bias enhancement. Additionally, higher graph connection density $\rho_d$, larger node number $n$, and balanced sensitive attribute ratio $c$ corresponds higher bias enhancement, which is consistent to our theoretical analysis in Theorem~\ref{theo:enhance}.  

\subsection{Experiments on Real-World Datasets}
\subsubsection{Experimental Settings}
\paragraph{Datasets.} We conduct experiments on real-world datasets Pokec-z, Pokec-n, and NBA \cite{dai2021say}. Pokec-z and Pokec-n are sampled, based on province information, from a larger Facebook-like social network Pokec \cite{takac2012data} in Slovakia, where region information is treated as the sensitive attribute and the predicted label is the working field of the users. NBA dataset is extended from a Kaggle dataset \footnote{https://www.kaggle.com/noahgift/social-power-nba} consisting of around 400 NBA basketball players. The information of players includes age, nationality, and salary in the 2016-2017 season. The players' link relationships is from Twitter with the official crawling API. The binary nationality (U.S. and overseas player) is adopted as the sensitive attribute and the prediction label is whether the salary is higher than the median. 

\paragraph{Evaluation Metrics.} We adopt accuracy to evaluate the performance of node classification task. As for fairness metric, we adopt two quantitative group fairness metrics to measure the prediction bias. According to works \cite{louizos2015variational,beutel2017data}, we adopt \emph{demographic parity} $\Delta_{DP}=|\mathbb{P}(\hat{y}=1|s=-1)-\mathbb{P}(\hat{y}=1|s=1)|$ and \emph{equal opportunity} $\Delta_{DP}=|\mathbb{P}(\hat{y}=1|s=-1, y=1)-\mathbb{P}(\hat{y}=1|s=1, y=1)|$, where $y$ and $\hat{y}$ represent the ground-truth label and predicted label, respectively. 

\paragraph{Baselines.} We compare our proposed FMP with representative GNNs, such as GCN \cite{kipf2017semi}, GAT \cite{velivckovic2018graph}, SGC \cite{wu2019simplifying}, and APPNP \cite{klicpera2019predict}, and MLP. For all models, we train 2 layers neural networks with 64 hidden units for $300$ epochs. Additionally, We also compare adversarial debiasing and adding demographic regularization methods to show the effectiveness of the proposed method.

\paragraph{Implementation Details.} We run the experiments $5$ times and report the average performance for each method. We adopt Adam optimizer with $0.001$ learning rate and $1e^{-5}$ weight decay for all models.
For adversarial debiasing, we adopt train classifier and adversary with $70$ and $30$ epochs, respectively. 
The hyperprameter for adversary loss is tuned in $\{0.0, 1.0, 2.0, 5.0, 8.0, 10.0, 20.0, 30.0\}$. For adding regularization, we adopt the hyperparameter set $\{0.0, 1.0, 2.0, 5.0, 8.0, 10.0, 20.0, 50.0, 80.0, 100.0\}$.

\begin{figure*}[t]
\centering
\includegraphics[width=0.85\linewidth]{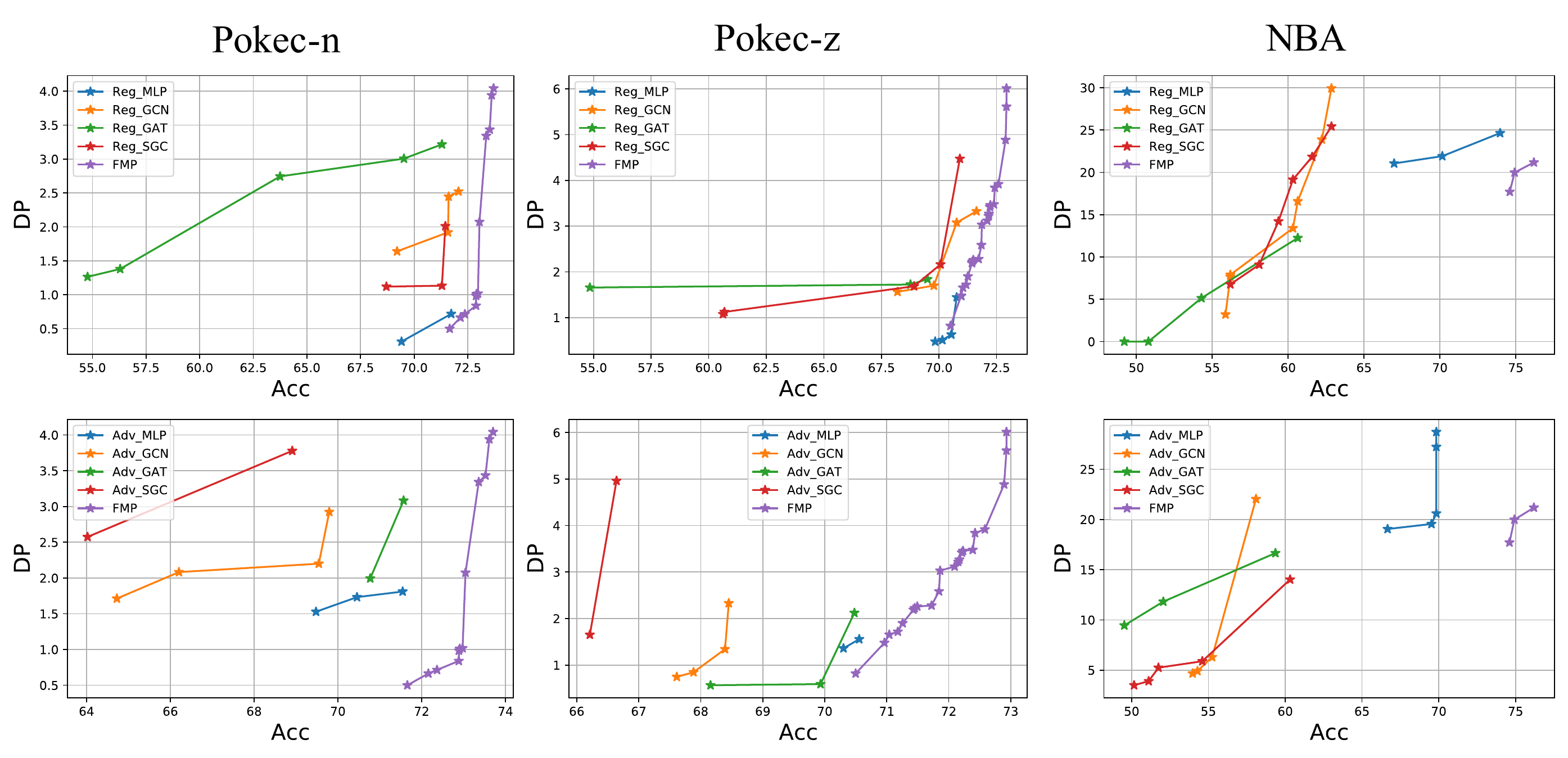}

\caption{DP and Acc trade-off performance on three real-world datasets compared with adding regularization (Top) and adversarial debiasing (Bottom). The trade-off curve more close to right bottom corner means better trade-off performance.}
\vspace{-15pt}
\label{fig:pareto}
\end{figure*}

\subsubsection{Experimental Results}
\paragraph{Comparison with existng GNNs.} The accuracy, demographic parity and equal opportunity metrics of proposed FMP for Pokec-z, Pokec-n, NBA dataset are shown in Table~\ref{table:comp_gnns} compared with MLP, GAT, GCN, SGC and APPNP. The detailed statistical information for these three dataset is shown in Table ~\ref{table:statistics}. From these results, we can obtain the following observations:
\begin{itemize}[leftmargin=0.2cm, itemindent=.0cm, itemsep=0.0cm, topsep=0.0cm]
    \item Many existing GNNs underperorm MLP model on all three datasets in terms of fairness metric. For instance, the demographic parity of MLP is lower than GAT, GCN, SGC and APPNP by $32.64\%$, $50.46\%$, $66.53\%$ and $58.72\%$ on Pokec-z dataset. The higher prediction bias comes from the aggregation within the-same-sensitive-attribute nodes and topology bias in graph data.
    \item Our proposed FMP consistently achieve lowest prediction bias in terms of demographic parity and equal opportunity on all datasets. Specifically, FMP reduces demographic parity by $49.69\%$, $56.86\%$ and $5.97\%$ compared with the lowest bias among all baselines in Pokec-z, Pokec-n, and NBA dataset. Meanwhile, our proposed FMP achieves the best accuracy in NBA dataset, and comparable accuracy in Pokec-z and Pokec-n datasets. In a nutshell, proposed FMP can effectively mitigate prediction bias while preserving the prediction performance.
\end{itemize}

\paragraph{Comparison with adversarial debiasing and regularization.} To validate the effectiveness of proposed FMP, we also show the prediction performance and fairness metric trade-off compared with fairness-boosting methods, including adversarial debiasing \cite{fisher2020debiasing} and adding regularization \cite{chuang2020fair}. Similar to \cite{louppe2017learning}, the output of GNNs is the input of adversary and the goal of adversary is to predict the node sensitive attribute. We also adopt several backbones for these two methods, including MLP, GCN, GAT and SGC. We randomly split $50\%/25\%/25\%$ for training, validation and test dataset. Figure~\ref{fig:pareto} shows the pareto optimality curve for all methods, where right-bottom corner point represents the ideal performance (highest accuracy and lowest prediction bias). From the results, we list the following observations as follows:
\begin{itemize}[leftmargin=0.2cm, itemindent=.0cm, itemsep=0.0cm, topsep=0.0cm]
    \item Our proposed FMP can achieve better DP-Acc trade-off compared with adversarial debiasing and adding regularization for many GNNs and MLP. Such observation validates the effectiveness of the key idea in FMP: aggregation first and then debiasing. Additionally, FMP can reduce demographic parity with negligible performance cost due to transparent and efficient debiasing.
    \item Message passing in GNNs does matter. For adding regularization or adversarial debiasing, different GNNs embrace huge distinction, which implies that appropriate message passing manner potentially leads to better trade-off performance. Additionally, many GNNs underperforms MLP in low label homophily coefficient dataset, such as NBA. The rationale is that aggregation may not always bring benefit in terms of accuracy when the neighbors have low probability with the same label.
\end{itemize}

\section{Related Works}
Due to the page limit, we briefly review the existing work on \textbf{graph neural networks} and \textbf{fairness-aware learning on graphs.} A more comprehensive discussion can be found in \underline{Appendix~\ref{app:related}}. Existing GNNs can be roughly divided into spectral-based and spatial-based GNNs. Spectral-based GNNs provide graph convolution definition based on graph theory \cite{bruna2013spectral,defferrard2016convolutional,henaff2015deep}.  Spatial-based GNNs variant are popular due to explicit neighbors' information aggregation, including Graph con-volutional networks (GCN) \cite{kipf2017semi}, graph attention network (GAT) \cite{velivckovic2018graph}. The unified optimization framework are also provided to unify many existing message passing schemes \cite{ma2021unified}. As for fairness in graph data, many works have been developed to achieve fairness in machine learning community \cite{chuang2020fair}, including fair walk \cite{rahman2019fairwalk}, adversarial debiasin \cite{dai2021say}, Bayesian approach \cite{buyl2020debayes}, and contrastive learning \cite{agarwal2021towards}. However, aforementioned works ignore the role of message passing in terms of fairness. In this work, we theoretically and experimentally reveal that GNNs aggregation schemes enhance node representation bias under topology bias, and develop a efficient and transparent fair message passing scheme via utilizing useful information of neighbors and debiasing node representation simultaneously. 

\section{Conclusion}
In this work, we theoretically demonstrate that the message passing enhances node representations bias under large sensitive homophily coefficient, and reveal the role of other graph statistical information in terms of bias enhancement. Additionally, integrated in a unified optimization framework, we develop an efficient, effective and transparent FMP to learn fair node representations while preserving prediction performance. Experimental results on real-world datasets demonstrate the effectiveness and efficiency compared with state-of-the-art baseline in node classification tasks.



\bibliography{fmp}

\begin{thebibliography}{47}
\providecommand{\natexlab}[1]{#1}
\providecommand{\url}[1]{\texttt{#1}}
\expandafter\ifx\csname urlstyle\endcsname\relax
  \providecommand{\doi}[1]{doi: #1}\else
  \providecommand{\doi}{doi: \begingroup \urlstyle{rm}\Url}\fi

\bibitem[Agarwal et~al.(2021)Agarwal, Lakkaraju, and
  Zitnik]{agarwal2021towards}
Agarwal, C., Lakkaraju, H., and Zitnik, M.
\newblock Towards a unified framework for fair and stable graph representation
  learning.
\newblock \emph{arXiv preprint arXiv:2102.13186}, 2021.

\bibitem[Beutel et~al.(2017)Beutel, Chen, Zhao, and Chi]{beutel2017data}
Beutel, A., Chen, J., Zhao, Z., and Chi, E.~H.
\newblock Data decisions and theoretical implications when adversarially
  learning fair representations.
\newblock \emph{arXiv preprint arXiv:1707.00075}, 2017.

\bibitem[Bose \& Hamilton(2019)Bose and Hamilton]{bose2019compositional}
Bose, A. and Hamilton, W.
\newblock Compositional fairness constraints for graph embeddings.
\newblock In \emph{International Conference on Machine Learning}, pp.\
  715--724. PMLR, 2019.

\bibitem[Bruna et~al.(2013)Bruna, Zaremba, Szlam, and LeCun]{bruna2013spectral}
Bruna, J., Zaremba, W., Szlam, A., and LeCun, Y.
\newblock Spectral networks and locally connected networks on graphs.
\newblock \emph{arXiv preprint arXiv:1312.6203}, 2013.

\bibitem[Buyl \& De~Bie(2020)Buyl and De~Bie]{buyl2020debayes}
Buyl, M. and De~Bie, T.
\newblock Debayes: a bayesian method for debiasing network embeddings.
\newblock In \emph{International Conference on Machine Learning}, pp.\
  1220--1229. PMLR, 2020.

\bibitem[Chen et~al.(2013)Chen, Huang, and Zhang]{chen2013primal}
Chen, P., Huang, J., and Zhang, X.
\newblock A primal--dual fixed point algorithm for convex separable
  minimization with applications to image restoration.
\newblock \emph{Inverse Problems}, 29\penalty0 (2):\penalty0 025011, 2013.

\bibitem[Chien et~al.(2021)Chien, Peng, Li, and Milenkovic]{chien2021adaptive}
Chien, E., Peng, J., Li, P., and Milenkovic, O.
\newblock Adaptive universal generalized pagerank graph neural network.
\newblock In \emph{International Conference on Learning Representations}, 2021.

\bibitem[Chuang \& Mroueh(2020)Chuang and Mroueh]{chuang2020fair}
Chuang, C.-Y. and Mroueh, Y.
\newblock Fair mixup: Fairness via interpolation.
\newblock In \emph{International Conference on Learning Representations}, 2020.

\bibitem[Creager et~al.(2019)Creager, Madras, Jacobsen, Weis, Swersky, Pitassi,
  and Zemel]{creager2019flexibly}
Creager, E., Madras, D., Jacobsen, J.-H., Weis, M., Swersky, K., Pitassi, T.,
  and Zemel, R.
\newblock Flexibly fair representation learning by disentanglement.
\newblock In \emph{International conference on machine learning}, pp.\
  1436--1445. PMLR, 2019.

\bibitem[Dai \& Wang(2021)Dai and Wang]{dai2021say}
Dai, E. and Wang, S.
\newblock Say no to the discrimination: Learning fair graph neural networks
  with limited sensitive attribute information.
\newblock In \emph{Proceedings of the 14th ACM International Conference on Web
  Search and Data Mining}, pp.\  680--688, 2021.

\bibitem[Defferrard et~al.(2016)Defferrard, Bresson, and
  Vandergheynst]{defferrard2016convolutional}
Defferrard, M., Bresson, X., and Vandergheynst, P.
\newblock Convolutional neural networks on graphs with fast localized spectral
  filtering.
\newblock \emph{Advances in neural information processing systems},
  29:\penalty0 3844--3852, 2016.

\bibitem[Dong et~al.(2016)Dong, Lizardo, and Chawla]{dong2016young}
Dong, Y., Lizardo, O., and Chawla, N.~V.
\newblock Do the young live in a “smaller world” than the old? age-specific
  degrees of separation in a large-scale mobile communication network.
\newblock \emph{arXiv preprint arXiv:1606.07556}, 2016.

\bibitem[Du et~al.(2021)Du, Mukherjee, Wang, Tang, Awadallah, and
  Hu]{du2021fairness}
Du, M., Mukherjee, S., Wang, G., Tang, R., Awadallah, A.~H., and Hu, X.
\newblock Fairness via representation neutralization.
\newblock \emph{arXiv preprint arXiv:2106.12674}, 2021.

\bibitem[Feldman et~al.(2015)Feldman, Friedler, Moeller, Scheidegger, and
  Venkatasubramanian]{feldman2015certifying}
Feldman, M., Friedler, S.~A., Moeller, J., Scheidegger, C., and
  Venkatasubramanian, S.
\newblock Certifying and removing disparate impact.
\newblock In \emph{proceedings of the 21th ACM SIGKDD international conference
  on knowledge discovery and data mining}, pp.\  259--268, 2015.

\bibitem[Fisher et~al.(2020)Fisher, Mittal, Palfrey, and
  Christodoulopoulos]{fisher2020debiasing}
Fisher, J., Mittal, A., Palfrey, D., and Christodoulopoulos, C.
\newblock Debiasing knowledge graph embeddings.
\newblock In \emph{Proceedings of the 2020 Conference on Empirical Methods in
  Natural Language Processing (EMNLP)}, pp.\  7332--7345, 2020.

\bibitem[Gao et~al.(2018)Gao, Wang, and Ji]{gao2018large}
Gao, H., Wang, Z., and Ji, S.
\newblock Large-scale learnable graph convolutional networks.
\newblock In \emph{Proceedings of the 24th ACM SIGKDD International Conference
  on Knowledge Discovery \& Data Mining}, pp.\  1416--1424, 2018.

\bibitem[Hamaguchi et~al.(2017)Hamaguchi, Oiwa, Shimbo, and
  Matsumoto]{hamaguchi2017knowledge}
Hamaguchi, T., Oiwa, H., Shimbo, M., and Matsumoto, Y.
\newblock Knowledge transfer for out-of-knowledge-base entities: A graph neural
  network approach.
\newblock \emph{arXiv preprint arXiv:1706.05674}, 2017.

\bibitem[Hamilton et~al.(2017)Hamilton, Ying, and
  Leskovec]{hamilton2017inductive}
Hamilton, W.~L., Ying, R., and Leskovec, J.
\newblock Inductive representation learning on large graphs.
\newblock In \emph{Proceedings of the 31st International Conference on Neural
  Information Processing Systems}, pp.\  1025--1035, 2017.

\bibitem[Henaff et~al.(2015)Henaff, Bruna, and LeCun]{henaff2015deep}
Henaff, M., Bruna, J., and LeCun, Y.
\newblock Deep convolutional networks on graph-structured data.
\newblock \emph{arXiv preprint arXiv:1506.05163}, 2015.

\bibitem[Kipf \& Welling(2017)Kipf and Welling]{kipf2017semi}
Kipf, T.~N. and Welling, M.
\newblock Semi-supervised classification with graph convolutional networks.
\newblock In \emph{International Conference on Learning Representations
  (ICLR)}, 2017.

\bibitem[Klicpera et~al.(2019)Klicpera, Bojchevski, and
  G{\"u}nnemann]{klicpera2019predict}
Klicpera, J., Bojchevski, A., and G{\"u}nnemann, S.
\newblock Predict then propagate: Graph neural networks meet personalized
  pagerank.
\newblock In \emph{International Conference on Learning Representations}, 2019.

\bibitem[K{\"o}se \& Shen(2021)K{\"o}se and Shen]{kose2021fairness}
K{\"o}se, {\"O}.~D. and Shen, Y.
\newblock Fairness-aware node representation learning.
\newblock \emph{arXiv preprint arXiv:2106.05391}, 2021.

\bibitem[Kullback(1997)]{kullback1997information}
Kullback, S.
\newblock \emph{Information theory and statistics}.
\newblock Courier Corporation, 1997.

\bibitem[Laclau et~al.(2021)Laclau, Redko, Choudhary, and
  Largeron]{laclau2021all}
Laclau, C., Redko, I., Choudhary, M., and Largeron, C.
\newblock All of the fairness for edge prediction with optimal transport.
\newblock In \emph{International Conference on Artificial Intelligence and
  Statistics}, pp.\  1774--1782. PMLR, 2021.

\bibitem[Li et~al.(2021)Li, Wang, Zhao, Hong, and Liu]{li2021dyadic}
Li, P., Wang, Y., Zhao, H., Hong, P., and Liu, H.
\newblock On dyadic fairness: Exploring and mitigating bias in graph
  connections.
\newblock In \emph{International Conference on Learning Representations}, 2021.

\bibitem[Liu et~al.(2021)Liu, Jin, Ma, Li, Liu, Wang, Yan, and
  Tang]{liu2021elastic}
Liu, X., Jin, W., Ma, Y., Li, Y., Liu, H., Wang, Y., Yan, M., and Tang, J.
\newblock Elastic graph neural networks.
\newblock In \emph{International Conference on Machine Learning}, pp.\
  6837--6849. PMLR, 2021.

\bibitem[Loris \& Verhoeven(2011)Loris and Verhoeven]{loris2011generalization}
Loris, I. and Verhoeven, C.
\newblock On a generalization of the iterative soft-thresholding algorithm for
  the case of non-separable penalty.
\newblock \emph{Inverse Problems}, 27\penalty0 (12):\penalty0 125007, 2011.

\bibitem[Louizos et~al.(2015)Louizos, Swersky, Li, Welling, and
  Zemel]{louizos2015variational}
Louizos, C., Swersky, K., Li, Y., Welling, M., and Zemel, R.
\newblock The variational fair autoencoder.
\newblock \emph{arXiv preprint arXiv:1511.00830}, 2015.

\bibitem[Louppe et~al.(2017)Louppe, Kagan, and Cranmer]{louppe2017learning}
Louppe, G., Kagan, M., and Cranmer, K.
\newblock Learning to pivot with adversarial networks.
\newblock In \emph{Proceedings of the 31st International Conference on Neural
  Information Processing Systems}, pp.\  982--991, 2017.

\bibitem[Ma et~al.(2021{\natexlab{a}})Ma, Deng, and Mei]{ma2021subgroup}
Ma, J., Deng, J., and Mei, Q.
\newblock Subgroup generalization and fairness of graph neural networks.
\newblock \emph{Advances in Neural Information Processing Systems}, 34,
  2021{\natexlab{a}}.

\bibitem[Ma et~al.(2021{\natexlab{b}})Ma, Liu, Shah, and Tang]{ma2021homophily}
Ma, Y., Liu, X., Shah, N., and Tang, J.
\newblock Is homophily a necessity for graph neural networks?
\newblock \emph{arXiv preprint arXiv:2106.06134}, 2021{\natexlab{b}}.

\bibitem[Ma et~al.(2021{\natexlab{c}})Ma, Liu, Zhao, Liu, Tang, and
  Shah]{ma2021unified}
Ma, Y., Liu, X., Zhao, T., Liu, Y., Tang, J., and Shah, N.
\newblock A unified view on graph neural networks as graph signal denoising.
\newblock In \emph{Proceedings of the 30th ACM International Conference on
  Information \& Knowledge Management}, pp.\  1202--1211, 2021{\natexlab{c}}.

\bibitem[Mehrabi et~al.(2021)Mehrabi, Morstatter, Saxena, Lerman, and
  Galstyan]{mehrabi2021survey}
Mehrabi, N., Morstatter, F., Saxena, N., Lerman, K., and Galstyan, A.
\newblock A survey on bias and fairness in machine learning.
\newblock \emph{ACM Computing Surveys (CSUR)}, 54\penalty0 (6):\penalty0 1--35,
  2021.

\bibitem[Monti et~al.(2017)Monti, Boscaini, Masci, Rodola, Svoboda, and
  Bronstein]{monti2017geometric}
Monti, F., Boscaini, D., Masci, J., Rodola, E., Svoboda, J., and Bronstein,
  M.~M.
\newblock Geometric deep learning on graphs and manifolds using mixture model
  cnns.
\newblock In \emph{Proceedings of the IEEE conference on computer vision and
  pattern recognition}, pp.\  5115--5124, 2017.

\bibitem[Rahman et~al.(2019)Rahman, Surma, Backes, and
  Zhang]{rahman2019fairwalk}
Rahman, T., Surma, B., Backes, M., and Zhang, Y.
\newblock Fairwalk: Towards fair graph embedding.
\newblock 2019.

\bibitem[Rockafellar(2015)]{rockafellar2015convex}
Rockafellar, R.~T.
\newblock \emph{Convex analysis}.
\newblock Princeton university press, 2015.

\bibitem[Suresh \& Guttag(2019)Suresh and Guttag]{suresh2019framework}
Suresh, H. and Guttag, J.~V.
\newblock A framework for understanding unintended consequences of machine
  learning.
\newblock \emph{arXiv preprint arXiv:1901.10002}, 2, 2019.

\bibitem[Takac \& Zabovsky(2012)Takac and Zabovsky]{takac2012data}
Takac, L. and Zabovsky, M.
\newblock Data analysis in public social networks.
\newblock In \emph{International scientific conference and international
  workshop present day trends of innovations}, volume~1, 2012.

\bibitem[Tang et~al.(2020)Tang, Yao, Sun, Wang, Tang, Aggarwal, Mitra, and
  Wang]{tang2020investigating}
Tang, X., Yao, H., Sun, Y., Wang, Y., Tang, J., Aggarwal, C., Mitra, P., and
  Wang, S.
\newblock Investigating and mitigating degree-related biases in graph
  convoltuional networks.
\newblock In \emph{Proceedings of the 29th ACM International Conference on
  Information \& Knowledge Management}, pp.\  1435--1444, 2020.

\bibitem[Veli{\v{c}}kovi{\'c} et~al.(2018)Veli{\v{c}}kovi{\'c}, Cucurull,
  Casanova, Romero, Li{\`o}, and Bengio]{velivckovic2018graph}
Veli{\v{c}}kovi{\'c}, P., Cucurull, G., Casanova, A., Romero, A., Li{\`o}, P.,
  and Bengio, Y.
\newblock Graph attention networks.
\newblock In \emph{International Conference on Learning Representations}, 2018.

\bibitem[Wu et~al.(2019)Wu, Souza, Zhang, Fifty, Yu, and
  Weinberger]{wu2019simplifying}
Wu, F., Souza, A., Zhang, T., Fifty, C., Yu, T., and Weinberger, K.
\newblock Simplifying graph convolutional networks.
\newblock In \emph{International conference on machine learning}, pp.\
  6861--6871. PMLR, 2019.

\bibitem[Ying et~al.(2018)Ying, He, Chen, Eksombatchai, Hamilton, and
  Leskovec]{ying2018graph}
Ying, R., He, R., Chen, K., Eksombatchai, P., Hamilton, W.~L., and Leskovec, J.
\newblock Graph convolutional neural networks for web-scale recommender
  systems.
\newblock In \emph{Proceedings of the 24th ACM SIGKDD International Conference
  on Knowledge Discovery \& Data Mining}, pp.\  974--983, 2018.

\bibitem[Yurochkin \& Sun(2020)Yurochkin and Sun]{yurochkin2020sensei}
Yurochkin, M. and Sun, Y.
\newblock Sensei: Sensitive set invariance for enforcing individual fairness.
\newblock In \emph{International Conference on Learning Representations}, 2020.

\bibitem[Zhang et~al.(2018)Zhang, Lemoine, and Mitchell]{zhang2018mitigating}
Zhang, B.~H., Lemoine, B., and Mitchell, M.
\newblock Mitigating unwanted biases with adversarial learning.
\newblock In \emph{Proceedings of the 2018 AAAI/ACM Conference on AI, Ethics,
  and Society}, pp.\  335--340, 2018.

\bibitem[Zhao \& Akoglu(2019)Zhao and Akoglu]{zhao2019pairnorm}
Zhao, L. and Akoglu, L.
\newblock Pairnorm: Tackling oversmoothing in gnns.
\newblock \emph{arXiv preprint arXiv:1909.12223}, 2019.

\bibitem[Zhu et~al.(2020)Zhu, Yan, Zhao, Heimann, Akoglu, and
  Koutra]{zhu2020beyond}
Zhu, J., Yan, Y., Zhao, L., Heimann, M., Akoglu, L., and Koutra, D.
\newblock Beyond homophily in graph neural networks: Current limitations and
  effective designs.
\newblock \emph{Advances in neural information processing systems}, 2020.

\bibitem[Zhu et~al.(2021)Zhu, Wang, Shi, Ji, and Cui]{zhu2021interpreting}
Zhu, M., Wang, X., Shi, C., Ji, H., and Cui, P.
\newblock Interpreting and unifying graph neural networks with an optimization
  framework.
\newblock In \emph{Proceedings of the Web Conference 2021}, pp.\  1215--1226,
  2021.

\end{thebibliography}
\bibliographystyle{icml2022}


\newpage
\appendix
\onecolumn
\section{Notations}
\begin{table}[h]
\caption{Table of Notations}
    \begin{tabular}{ cc} 
    \toprule
    Notations & Description \\
    \hline
    $|\mathcal{E}|$ & The number of edges \\
    $n, d, d_{out}$ & The number of nodes; the number of node feature dimensions; the number of node classes\\
    $\Delta_{\mathbf{s}}\in\mathbb{R}^{1\times n}$ & The sensitive attribute incident vector \\
    $\epsilon_{label}$ & Label homophily coefficient \\
    $\epsilon_{sens}$ & Sensitive homophily coefficient \\
    $\mathbf{X}_{ori}\in\mathbb{R}^{n\times d}$ & The input node attributes matrix \\
    $\mathbf{A}\in\mathbb{R}^{n\times n}$ & The adjacency matrix\\
    $\hat{\mathbf{A}}\in\mathbb{R}^{n\times n}$ & The adjacency matrix with self-loop\\
    $\tilde{\mathbf{A}}\in\mathbb{R}^{n\times n}$ & The normalized adjacency matrix with self-loop\\
    $\mathbf{L}\in\mathbb{R}^{n\times n}$ & The Laplacian matrix\\
    $\mathbf{X}_{trans}\in\mathbb{R}^{n\times d_{out}}$ & The output node features for feature transformation \\
    $\mathbf{F}_{agg}\in\mathbb{R}^{n\times d_{out}}$ & The aggregated node features after propagation \\
    $\mathbf{F}\in\mathbb{R}^{n\times d_{out}}$ & The learned node features considering graph smoothness and fairness \\
    $\mathbf{u}\in\mathbb{R}^{1\times d_{out}}$ & The permutation direction in feature representation space \\
    $h^{*}(\cdot)$ & Fenchel conjugate function of $h(\cdot)$ \\
    $||\mathbf{X}||_F$, $||\mathbf{X}||_1$ & The Frobenius norm and $l_1$ norm of matrix $\mathbf{X}$\\
    $\lambda_f$, $\lambda_s$ & Hyperparameter for fairness and graph smoothness objectives \\
    \bottomrule 
    \end{tabular}
\end{table}

\section{Proof of Theorem~\ref{theo:surro}}\label{app:surro}
We provide a more general proof for categorical sensitive attribute $\mathbf{s}\in\{1,2,\cdots, K\}$ and the prior probability is given by $\mathbb{P}(\mathbf{s}=i)=c_i$. Suppose the conditional node attribute $\mathbf{x}$ distribution given node sensitive attribute $\mathbf{s}=i$ satisfies normal distribution $P_i(\mathbf{x})\dff\mathcal{N}(\mathbf{\mu}_i, \mathbf{\Sigma_i})$, the distribution of node sensitive attribute is the mixed Gaussian distribution $f_{}(\mathbf{x})=\sum_{i=1}^{K}c_iP_i(\mathbf{x})$.
Based on the definition of mutual information, we have
\be 
I(\mathbf{s}, \mathbf{x})=H(\mathbf{x}) - \sum_{i=1}^{K}c_iH(\mathbf{x}|\mathbf{s}=i);
\ee 
where $H(\cdot)$ represents Shannon entropy for random variable. Subsequently, we focus on the entropy of the mixed Gaussian distribution $H(\mathbf{x})$. We show that such entropy can be upper bounded by the pairwise Kullback-Leibler (KL) divergence as follows:
\be 
I(\mathbf{s}, \mathbf{x}) &=& -\sum_{i=1}^{K}c_i\mathbb{E}_{P_i}\big[\ln\sum_{j=1}^{K}c_jP_j(\mathbf{x})\big]-\sum_{i=1}^{K}c_iH(\mathbf{x}|\mathbf{s}=i)\nonumber\\
&\overset{(a)}{\leq}&-\sum_{i=1}^{K}c_i\big[\ln\sum_{j=1}^{K}c_j e^{\mathbb{E}_{P_i}[\ln P_j(\mathbf{x})]}\big]-\sum_{i=1}^{K}c_iH(\mathbf{x}|\mathbf{s}=i)\nonumber\\
&=&-\sum_{i=1}^{K}c_i\big[\ln\sum_{j=1}^{K}c_j e^{-H(P_i||P_j)}\big]-\sum_{i=1}^{K}c_iH(\mathbf{x}|\mathbf{s}=i)\nonumber\\
&\overset{(b)}{=}&-\sum_{i=1}^{K}c_i\big[\ln\sum_{j=1}^{K}c_j e^{-H(P_i)-D_{KL}(P_i||P_j)}\big]-\sum_{i=1}^{K}c_iH(\mathbf{x}|\mathbf{s}=i)\nonumber\\
&=& -\sum_{i=1}^{K}c_i\big[\ln\sum_{j=1}^{K}c_j e^{-D_{KL}(P_i||P_j)}\big], \nonumber
\ee 
where KL divergence $D_{KL}(P_i||P_j)\dff \int P_i(\mathbf{x})\ln\frac{P_i(\mathbf{x})}{P_j(\mathbf{x})}\mathrm{d}\mathbf{x}$ and cross entropy $H(P_i||P_j)=-\int P_i(\mathbf{x})\ln P_j(\mathbf{x})\mathrm{d}\mathbf{x}$. The inequality (a) holds based on the variational lower bound on the expectation of a log-sum inequality $\mathbb{E}\big[\ln\sum_i Z_i\big]\geq \ln\big[\sum_i e^{\mathbb{E}[\ln Z_i]}\big]$ \cite{kullback1997information}, and quality $(2)$ holds based on $H(P_i||P_j)=H(P_i)+D_{KL}(P_i||P_j)$. As a special case for binary sensitive attribute, it is easy to obtain the following results:
\be 
I(\mathbf{s}, \mathbf{X})\leq -(1-c)\ln\Big[(1-c) +c\exp\big(-D_{KL}(P_1||P_2)\big)\Big]
-c\ln\Big[c +(1-c)\exp\big(-D_{KL}(P_2||P_1)\big)\Big]. \nonumber
\ee 

\section{Proof of Theorem~\ref{theo:enhance}} \label{app:enhance}
Before going deeper for our proof, we first introduce two useful lemmas on KL divergence and statistical information of graph.
\begin{lemma}\label{lemma:KL}
For two $d$-dimensional Gaussian distributions $P=\mathcal{N}(\mathbf{\mu}_p, \mathbf{\Sigma}_p)$ and $Q=\mathcal{N}(\mathbf{\mu}_q, \mathbf{\Sigma}_q)$, the KL divergence $D_{KL}(P||Q)$ is given by
\be 
D_{KL}(P||Q)=\frac{1}{2}\Big[\ln\frac{|\mathbf{\Sigma}_q|}{|\mathbf{\Sigma}_p|}-d+(\mathbf{\mu}_p-\mathbf{\mu}_q)^{\top}\mathbf{\Sigma}_q^{-1}(\mathbf{\mu}_p-\mathbf{\mu}_q)+Tr(\mathbf{\Sigma}_q^{-1}\mathbf{\Sigma}_p)\Big]
\ee 
where $\top$ is matrix transpose operation and $Tr(\cdot)$ is trace of a square matrix.
\begin{proof}
Note that the probability density function of multivariate Normal distribution is given by: 
\be 
P(\mathbf{x})=\frac{1}{(2\pi)^{d/2}|\mathbf{\Sigma}_p|^{1/2}}\exp\Big(-\frac{1}{2}(\mathbf{x}-\mathbf{\mu}_p)^{\top}\mathbf{\Sigma}_p^{-1}(\mathbf{x}-\mathbf{\mu}_p)\Big), \nonumber
\ee 
the KL divergence between distributions $P$ and $Q$ can be given by 
\be 
D_{KL}(P||Q)&=&\mathbb{E}_{P}[\ln(P)-\ln(Q)] \nonumber\\
&=&\mathbb{E}_{P}\Big[\frac{1}{2}\ln\frac{|\mathbf{\Sigma}_q|}{|\mathbf{\Sigma}_p|}-\frac{1}{2}(\mathbf{x}-\mathbf{\mu}_p)^{\top}\mathbf{\Sigma}_p^{-1}(\mathbf{x}-\mathbf{\mu}_p)+\frac{1}{2}(\mathbf{x}-\mathbf{\mu}_q)^{\top}\mathbf{\Sigma}_q^{-1}(\mathbf{x}-\mathbf{\mu}_q)\Big]\nonumber\\
&=&\frac{1}{2}\ln\frac{|\mathbf{\Sigma}_q|}{|\mathbf{\Sigma}_p|}-\underbrace{\frac{1}{2}\mathbb{E}_{P}\Big[(\mathbf{x}-\mathbf{\mu}_p)^{\top}\mathbf{\Sigma}_p^{-1}(\mathbf{x}-\mathbf{\mu}_p)\Big]}_{I_1}+\underbrace{\frac{1}{2}\mathbb{E}_P\Big[(\mathbf{x}-\mathbf{\mu}_q)^{\top}\mathbf{\Sigma}_q^{-1}(\mathbf{x}-\mathbf{\mu}_q)\Big]}_{I_2}.
\ee
Using the commutative property of the trace operation, we have 
\be 
I_1 &=& \frac{1}{2}\mathbb{E}_{P}\Big[(\mathbf{x}-\mathbf{\mu}_p)^{\top}\mathbf{\Sigma}_p^{-1}(\mathbf{x}-\mathbf{\mu}_p)\Big] = \frac{1}{2}Tr\Big(\mathbb{E}_{P}\Big[(\mathbf{x}-\mathbf{\mu}_p)^{\top}(\mathbf{x}-\mathbf{\mu}_p)\mathbf{\Sigma}_p^{-1}\Big]\Big) \nonumber\\
&=&\frac{1}{2}Tr\Big(\mathbb{E}_{P}\Big[(\mathbf{x}-\mathbf{\mu}_p)^{\top}(\mathbf{x}-\mathbf{\mu}_p)\Big]\mathbf{\Sigma}_p^{-1}\Big)=\frac{1}{2}Tr\Big(\mathbf{\Sigma}_p\mathbf{\Sigma}_p^{-1}\Big)=\frac{d}{2},
\ee 
As for the term $I_2$, note that $\mathbf{x}-\mathbf{\mu}_q=(\mathbf{x}-\mathbb{E}_P[\mathbf{x}])+(\mathbb{E}_P[\mathbf{x}]-\mathbf{\mu}_q)$, we can obtain the following equation:
\be 
I_2&=&\frac{1}{2}\mathbb{E}_P\Big[(\mathbf{x}-\mathbf{\mu}_q)^{\top}\mathbf{\Sigma}_q^{-1}(\mathbf{x}-\mathbf{\mu}_q)\Big]\nonumber\\
&=&\frac{1}{2}\Big[(\mathbf{\mu}_p-\mathbf{\mu}_q)^{\top}\mathbf{\Sigma}_q^{-1}(\mathbf{\mu}_p-\mathbf{\mu}_q)+Tr(\mathbf{\Sigma}_q^{-1}\mathbf{\Sigma}_p)\Big],
\ee 
Therefore, the KL divergence $D_{KL}(P||Q)$ is given by
\be 
D_{KL}(P||Q)=\frac{1}{2}\Big[\ln\frac{|\mathbf{\Sigma}_q|}{|\mathbf{\Sigma}_p|}-d+(\mathbf{\mu}_p-\mathbf{\mu}_q)^{\top}\mathbf{\Sigma}_q^{-1}(\mathbf{\mu}_p-\mathbf{\mu}_q)+Tr(\mathbf{\Sigma}_q^{-1}\mathbf{\Sigma}_p)\Big].
\ee 
\end{proof}
\end{lemma}

\begin{lemma}\label{lemma:statistical}
Suppose the synthetic graph is generated from $(n, \rho_d,\epsilon_{sens}, c)$-graph, then we obtain the intra-connect and inter-connect probability as follows:
\be 
p_{conn}=\frac{\rho_d \epsilon_{sens}}{c^2+(1-c)^2}, \quad
q_{conn}=\frac{\rho_d (1-\epsilon_{sens})}{2c(1-c)}\nonumber.
\ee 
\begin{proof}
Based on Bayes' rule, we have  the intra-connect and inter-connect probability as follows
\be 
p_{conn}=\mathbb{P}(\mathbf{A}_{ij}=1|\mathbf{s}_i=\mathbf{s}_j)=\frac{\mathbb{P}(\mathbf{A}_{ij}=1)\mathbb{P}(\mathbf{s}_i=\mathbf{s}_j|\mathbf{A}_{ij}=1)}{\mathbb{P}(\mathbf{s}_i=\mathbf{s}_j)}=\frac{\rho_d \epsilon_{sens}}{c^2+(1-c)^2}, \nonumber\\
q_{conn}=\mathbb{P}(\mathbf{A}_{ij}=1|\mathbf{s}_i\mathbf{s}_j=-1)=\frac{\mathbb{P}(\mathbf{A}_{ij}=1)\mathbb{P}(\mathbf{s}_i\mathbf{s}_j=-1|\mathbf{A}_{ij}=1)}{\mathbb{P}(\mathbf{s}_i\mathbf{s}_j)=-1}=\frac{\rho_d (1-\epsilon_{sens})}{2c(1-c)}.
\ee 
\end{proof}
\end{lemma}
Note that the synthetic graph is generated from $(n, \rho_d, \epsilon_{sens}, c)$-graph. The sensitive attribute $\mathbf{s}$ is generated to with ratio $c$, i.e., the number of node sensitive attribute $\mathbf{s}=-1$ and $\mathbf{s}=1$ are $n_{-1}=n(1-c)$ and $n_{1}=nc$. Based on the determined sensitive attribute $\mathbf{s}$, we randomly generate the edge based on parameters $\rho_d$ and $\epsilon_{sens}$ and Lemma~\ref{lemma:statistical}, i.e., the edges within and cross the same group are randomly generated based on intra-connect probability and inter-connect probability. Therefore, the adjacency matrix $\mathbf{A}_{ij}$ is independent on node attribute $\mathbf{X}_i$ and $\mathbf{X}_j$ given sensitive attributes $\mathbf{s}_i$ and $\mathbf{s}_j$, i.e., $\mathbf{A}_{ij}\indep (\mathbf{X}_i, \mathbf{X}_j)|(\mathbf{s}_i, \mathbf{s}_j)$. Similarly, the different node attributes and edges are also dependent for each other given sensitive attributes, i.e., $\mathbf{A}_{ij}\indep \mathbf{A}_{ij} |(\mathbf{s}_i, \mathbf{s}_j, \mathbf{s}_k)$ and $\mathbf{X}_i\indep \mathbf{X}_j|(\mathbf{s}_i, \mathbf{s}_j)$.
Therefore, considering GCN-like message passing $\tilde{\mathbf{X}}_i=\sum_{j=1}^{n}\tilde{\mathbf{A}_{ij}}\mathbf{X}_j$, we have the aggregated node attributes expectation given sensitive attribute as follows:
\be 
\tilde{\mathbf{\mu}}_1&=&\mathbb{E}_{\tilde{\mathbf{X}}_i}[\tilde{\mathbf{X}}_i|\mathbf{s}_i=-1]=\sum_{j=1}^{n}\mathbb{E}_{\tilde{\mathbf{A}}_{ij},\mathbf{X}_j}[\tilde{\mathbf{A}}_{ij}\mathbf{X}_j|\mathbf{s}_i=-1]=\sum_{j=1}^{n}\mathbb{E}_{\tilde{\mathbf{A}}_{ij}}[\tilde{\mathbf{A}}_{ij}|\mathbf{s}_i=-1]\mathbf{E}_{\mathbf{X}_j}[\mathbf{X}_j|\mathbf{s}_i=-1]\nonumber\\
&=&(n_{-1}-1)\mathbb{E}_{\tilde{\mathbf{A}}_{ij}}[\tilde{\mathbf{A}}_{ij}|\mathbf{s}_i=-1, \mathbf{s}_j=-1]\mathbf{E}_{\mathbf{X}_j}[\tilde{\mathbf{X}}_j|\mathbf{s}_i=-1,\mathbf{s}_j=-1]\nonumber\\ 
&&+ \mathbb{E}_{\tilde{\mathbf{A}}_{ij}}[\tilde{\mathbf{A}}_{ij}|\mathbf{s}_i=-1, i=j]\mathbf{E}_{\mathbf{X}_j}[\mathbf{X}_j|\mathbf{s}_i=-1] \nonumber\\ 
&&+ n_{1} \mathbb{E}_{\tilde{\mathbf{A}}_{ij}}[\tilde{\mathbf{A}}_{ij}|\mathbf{s}_i=-1, \mathbf{s}_j=1]\mathbf{E}_{\mathbf{X}_j}[\mathbf{X}_j|\mathbf{s}_i=-1, \mathbf{s}_j=1]\nonumber\\
&=&\frac{[(n_{-1}-1)p_{conn}+1]\mathbf{\mu}_1+n_1 q_{conn}\mathbf{\mu}_2}{(n_{-1}-1)p_{conn}+1+n_1 q_{conn}}\dff \nu_1 \mathbf{\mu}_1 + (1-\nu_1)\mathbf{\mu}_2.
\ee 
where $\nu_1=\frac{(n_{-1}-1)p_{conn}+1}{(n_{-1}-1)p_{conn}+1+n_1 q_{conn}}$. Similarly, for the node with sensitive attribute $1$, we have
\be 
\tilde{\mathbf{\mu}}_2&=&\mathbb{E}_{\tilde{\mathbf{X}}_i}[\tilde{\mathbf{X}}_i|\mathbf{s}_i=1]=\sum_{j=1}^{n}\mathbb{E}_{\tilde{\mathbf{A}}_{ij},\mathbf{X}_j}[\tilde{\mathbf{A}}_{ij}\mathbf{X}_j|\mathbf{s}_i=1]=\sum_{j=1}^{n}\mathbb{E}_{\tilde{\mathbf{A}}_{ij}}[\tilde{\mathbf{A}}_{ij}|\mathbf{s}_i=1]\mathbf{E}_{\mathbf{X}_j}[\mathbf{X}_j|\mathbf{s}_i=1]\nonumber\\
&=&n_{-1}\mathbb{E}_{\tilde{\mathbf{A}}_{ij}}[\tilde{\mathbf{A}}_{ij}|\mathbf{s}_i=1, \mathbf{s}_j=-1]\mathbf{E}_{\mathbf{X}_j}[\mathbf{X}_j|\mathbf{s}_i=1,\mathbf{s}_j=-1]\nonumber\\ 
&&+ \mathbb{E}_{\tilde{\mathbf{A}}_{ij}}[\tilde{\mathbf{A}}_{ij}|\mathbf{s}_i=1, i=j]\mathbf{E}_{\mathbf{X}_j}[\mathbf{X}_j|\mathbf{s}_i=1] \nonumber\\ 
&&+ (n_{1}-1) \mathbb{E}_{\tilde{\mathbf{A}}_{ij}}[\tilde{\mathbf{A}}_{ij}|\mathbf{s}_i=1, \mathbf{s}_j=1]\mathbf{E}_{\mathbf{X}_j}[\mathbf{X}_j|\mathbf{s}_i=1, \mathbf{s}_j=1]\nonumber\\
&=&\frac{n_{-1}q_{conn}\mathbf{\mu}_1+[(n_1-1) p_{conn}+1]\mathbf{\mu}_2}{n_{-1}q_{conn}+(n_1-1) p_{conn}+1}\dff \nu_2 \mathbf{\mu}_1 + (1-\nu_2)\mathbf{\mu}_2.
\ee 
where $\nu_2=\frac{(n_{-1}-1)q_{conn}}{n_{-1}q_{conn}+1+(n_1-1) p_{conn}}$. 
As for the covariance matrix of aggregated node attributes $\tilde{\mathbf{X}}$ given node sensitive attribute $\mathbf{s}=-1$ and original sensitive attribute, note that 
we can obtain 
\be 
\tilde{\mathbf{\Sigma}}_1&=&\mathbb{D}_{\tilde{\mathbf{X}}_i}[\tilde{\mathbf{X}}_i|\mathbf{s}_i=-1]=\sum_{j=1}^{n}\mathbb{D}_{\tilde{\mathbf{A}}_{ij},\mathbf{X}_j}[\tilde{\mathbf{A}}_{ij}\mathbf{X}_j|\mathbf{s}_i=-1]=\sum_{j=1}^{n}\mathbb{E}_{\tilde{\mathbf{A}}_{ij}}[\tilde{\mathbf{A}}_{ij}^2|\mathbf{s}_i=-1]\mathbf{D}_{\tilde{\mathbf{X}}_j}[\tilde{\mathbf{X}}_j|\mathbf{s}_i=-1]\nonumber\\
&=& \frac{(n_{-1}-1)p_{conn}\mathbf{\Sigma}+\mathbf{\Sigma}+n_1 q_{conn}\mathbf{\Sigma}}{[(n_{-1}-1)p_{conn}+1+n_1 q_{conn}]^2}=\frac{\mathbf{\Sigma}}{(n_{-1}-1)p_{conn}+1+n_1 q_{conn}}\dff \zeta_1^{-1}\mathbf{\Sigma},
\ee 
where $\zeta_1=(n_{-1}-1)p_{conn}+1+n_1 q_{conn}$. 
Similarly, given node sensitive attribute $\mathbf{s}=1$, we have $\tilde{\mathbf{\Sigma}}_2=\mathbb{D}_{\tilde{\mathbf{X}}_i}[\tilde{\mathbf{X}}_i|\mathbf{s}_i=1]=\frac{\mathbf{\Sigma}}{n_{-1}q_{conn}+1+(n_1-1) p_{conn}}\dff \zeta_2^{-1}\mathbf{\Sigma}$, where $\zeta_2=n_{-1}q_{conn}+1+(n_1-1) p_{conn}$. In other words, the covariance matrix of the aggregated node attributes is lower than original one since ``average" operation will make node representation more concentrated. Note that the summation over several Gaussian random variables is still Gaussian, we define the node attributes distribution for sensitive attribute $\mathbf{s}=-1$ and $\mathbf{s}=1$ as $P_1=\mathcal{N}(\mathbf{\mu}_1, \mathbf{\Sigma})$, $P_2=\mathcal{N}(\mathbf{\mu}_2, \mathbf{\Sigma})$, respectively. Similarly, the aggregated node representation distribution follows for sensitive attribute $\mathbf{s}=-1$ and $\mathbf{s}=1$ as $\tilde{P}_1=\mathcal{N}(\tilde{\mathbf{\mu}}_1, \tilde{\mathbf{\Sigma}}_1)$, $\tilde{P}_2=\mathcal{N}(\tilde{\mathbf{\mu}}_2, \tilde{\mathbf{\Sigma}}_2)$. Note that the sensitive attribute ratio keeps the same after the aggregation and lager KL divergence for these two sensitive attributes group distribution, the bias enhances $\Delta Bias >0$ if $D_{KL}(\tilde{P}_1||\tilde{P}_2)>D_{KL}(P_1||P_2)$ and $D_{KL}(\tilde{P}_2||\tilde{P}_1)>D_{KL}(P_2||P_1)$. There fore, we only focus on the KL divergence. According to Lemma~\ref{lemma:KL}, it is easy to obtain KL divergence for original distribution as follows:
\be \label{eq:KL_ori}
D_{KL}(P_1||P_2)= \frac{1}{2}\Big[\ln\frac{|\mathbf{\Sigma}|}{|\mathbf{\Sigma}|}-d+(\mathbf{\mu}_1-\mathbf{\mu}_2)^{\top}\mathbf{\Sigma}^{-1}(\mathbf{\mu}_1-\mathbf{\mu}_2)+Tr(\mathbf{\Sigma}^{-1}\mathbf{\Sigma})\Big] = \frac{1}{2}(\mathbf{\mu}_1-\mathbf{\mu}_2)^{\top}\mathbf{\Sigma}^{-1}(\mathbf{\mu}_1-\mathbf{\mu}_2),
\ee 
As for KL divergence for aggregated distribution, similarly, we have 
\be \label{eq:KL_agg}
D_{KL}(\tilde{P}_1||\tilde{P}_2)&=& \frac{1}{2}\Big[\ln\frac{|\tilde{\mathbf{\Sigma}}_2|}{|\tilde{\mathbf{\Sigma}}_1|}-d+(\tilde{\mathbf{\mu}}_1-\tilde{\mathbf{\mu}}_2)^{\top}\tilde{\mathbf{\Sigma}}_2^{-1}(\tilde{\mathbf{\mu}}_1-\tilde{\mathbf{\mu}}_2)+Tr(\tilde{\mathbf{\Sigma}}_2^{-1}\tilde{\mathbf{\Sigma}}_1)\Big] \nonumber\\
&=&\frac{1}{2}\Big[d\ln\frac{\zeta_1}{\zeta_2} -d + (\nu_1-\nu_2)^2\zeta_2 (\mathbf{\mu}_1-\mathbf{\mu}_2)^{\top}\mathbf{\Sigma}^{-1}(\mathbf{\mu}_1-\mathbf{\mu}_2)+ \frac{\zeta_2}{\zeta_1}Tr(\mathbf{I}_d)\Big] \nonumber\\
&\overset{(c)}{\geq}& \frac{1}{2}(\nu_1-\nu_2)^2\zeta_2 (\mathbf{\mu}_1-\mathbf{\mu}_2)^{\top}\mathbf{\Sigma}^{-1}(\mathbf{\mu}_1-\mathbf{\mu}_2),
\ee  
where inequality $(c)$ holds since $\ln x\leq x-1$ for any $x>0$. Compared with equations (\ref{eq:KL_ori}) and (\ref{eq:KL_agg}), it is seen that $D_{KL}(\tilde{P}_1||\tilde{P}_2)>D_{KL}(P_1||P_2)$ if $(\nu_1-\nu_2)^2\zeta_2>1$. Similarly, we can have $D_{KL}(\tilde{P}_2||\tilde{P}_1)>D_{KL}(P_2||P_1)$ if $(\nu_1-\nu_2)^2\zeta_1>1$. In a nutshell, the bias enhances $\Delta Bias >0$ after message passing if $(\nu_1-\nu_2)^2\min\{\zeta_1, \zeta_2\}>1$.

\section{More Details on FMP Derivation}\label{app:fmp}
\subsection{Proof on Fairness Objective}\label{app:fairnessobj}
The fairness objective can be shown as the average prediction probability difference as follows:
\be 
\Big(\Delta_s SF(\mathbf{F})\big)\Big)_j&=& \Big[\frac{\mathbbm{1}_{>0}(\mathbf{s})}{||\mathbbm{1}_{>0}(\mathbf{s})||_1} - \frac{\mathbbm{1}_{>0}(-\mathbf{s})}{||\mathbbm{1}_{>0}(-\mathbf{s})||_1}\Big] \big(SF(\mathbf{F})\big)_{:,j}\nonumber\\
&=&\frac{\sum_{\mathbf{s}_i=1}\hat{P}(y_i=j|\mathbf{X})}{||\mathbbm{1}_{>0}(\mathbf{s})||_1}-\frac{\sum_{\mathbf{s}_i=-1}\hat{P}(y_i=j|\mathbf{X})}{||\mathbbm{1}_{>0}(\mathbf{-s})||_1}\nonumber\\
&=&\hat{P}(y_i=j|\mathbf{s}_i=1, \mathbf{X}) - \hat{P}(y_i=j|\mathbf{s}_i=-1, \mathbf{X}). \nonumber
\ee 
\subsection{Fenchel Conjugate}
Fenchel conjugate \cite{rockafellar2015convex} (a.k.a. convex conjugate) is the key tool to transform the original problem into bi-level optimization problem. For the general convex function $h(\cdot)$, its conjugate function is defined as 
\be 
h^{*}(\mathbf{U})\dff \sup\limits_{\mathbf{X}}\langle \mathbf{U}, \mathbf{X}\rangle -h(\mathbf{X}). \nonumber
\ee 
Based on Fenchel conjugate, the fairness objective can be transformed as variational representation $h_f(\mathbf{p})=\sup\limits_{\mathbf{u}}\langle\mathbf{p},\mathbf{u} \rangle - h_f^{*}(\mathbf{u})$, where $\mathbf{p}=\mathbf{\Delta}_s SF(\mathbf{F})\in\mathbb{R}^{1\times d_{out}}$ is a predicted probability vector for classification. Furthermore, the original optimization problem is equivalent to $\min\limits_{\mathbf{F}}\max\limits_{\mathbf{u}} h_s(\mathbf{F}) + \langle\mathbf{p},\mathbf{u} \rangle - h_f^{*}(\mathbf{u})$,
where $\mathbf{u}\in\mathbb{R}^{1\times d_{out}}$ and $h_f^{*}(\cdot)$ is the conjugate function of fairness objective $h_f(\cdot)$.

\subsection{Fair Message Passing Scheme.} 
Motivated by Proximal Alternating Predictor-Corrector (PAPC) \cite{loris2011generalization,chen2013primal}, the min-max optimization problem (\ref{eq:minmax}) can be solved by the following fixed-point equations
\be 
\left\{ 
\begin{array}{l}
     \mathbf{F}=\mathbf{F}-\nabla h_s(\mathbf{F})-\frac{\partial \langle \mathbf{p}, \mathbf{u}\rangle}{\partial \mathbf{F}}, \\
     \mathbf{u} = \text{prox}_{h^{*}_{f}}\big(\mathbf{u}+ \mathbf{\Delta_{s}} SF(\mathbf{F})\big).
\end{array}
\right.
\ee 
where $\text{prox}_{h^{*}_{f}}(\mathbf{u})=\arg\min\limits_{\mathbf{y}}||\mathbf{y}-\mathbf{u}||_F^2+h^{*}_{f}(\mathbf{u})$. Similar to ``predictor-corrector" algorithm \cite{loris2011generalization}, we adopt iterative algorithm to find saddle point for min-max optimization problem. Specifically, starting from $(\mathbf{F}^{k}, \mathbf{u}^{k})$, we adopt a gradient descent step on primal variable $\mathbf{F}$ to arrive $(\bar{\mathbf{F}}^{k+1}, \mathbf{u}^{k})$ and then followed by a proximal ascent step in the dual variable $\mathbf{u}$. Finally, a gradient descent step on primal variable in point $(\bar{\mathbf{F}}^{k+1}, \mathbf{u}^{k})$ to arrive at $(\mathbf{F}^{k+1}, \mathbf{u}^{k})$. In short, the iteration can be summarized as
\be 
\left\{ 
\begin{array}{l}
     \bar{\mathbf{F}}^{k+1}=\mathbf{F}^{k}-\gamma\nabla h_s(\mathbf{F}^{k})-\gamma\frac{\partial \langle \mathbf{p}, \mathbf{u}^{k}\rangle}{\partial \mathbf{F}}\Big|_{\mathbf{F}^{k}}, \\
     \mathbf{u}^{k+1} = \text{prox}_{\beta h^{*}_{f}}\big(\mathbf{u}^{k}+\beta \mathbf{\Delta_{s}} SF(\bar{\mathbf{F}}^{k+1})\big), \\
     \bar{\mathbf{F}}^{k+1}=\mathbf{F}^{k}-\gamma\nabla h_s(\mathbf{F}^{k})-\gamma\frac{\partial \langle \mathbf{p}, \mathbf{u}^{k+1}\rangle}{\partial \mathbf{F}}\Big|_{\mathbf{F}^{k}}. \\
\end{array}
\right.
\ee 
where $\gamma$ and $\beta$ are the step size for primal and dual variables. Note that the close-form for $\frac{\partial \langle \mathbf{p}, \mathbf{u}\rangle}{\partial \mathbf{F}}\in\mathbb{R}^{n\times d_{out}}$ and $\text{prox}_{\beta h^{*}_{f}}(\cdot)$ are still not clear, we will provide the solution on by one.

\paragraph{Proximal Operator.} As for the proximal operators, we provide the close-form in the following proposition:
\begin{proposition}[Proximal Operators]\label{prop:conjugate}
The proximal operators $\text{prox}_{\beta h^{*}_{f}}(\mathbf{u})$ satisfies 
\be 
\text{prox}_{\beta h^{*}_{f}}(\mathbf{u})_{j}=sign(\mathbf{u})_{j}\min\big(|\mathbf{u}_{j}|, \lambda_f\big),
\ee 
where $sign(\cdot)$ and $\lambda_f$ are element-wise sign function and hyperparameter for fairness objective. In other words, such proximal operator is element-wise projection onto $l_{\infty}$ ball with radius $\lambda_f$.
\end{proposition}

\paragraph{FMP Scheme.} Similar to works \cite{ma2021unified,liu2021elastic}, choosing $\gamma=\frac{1}{1+\lambda_f}$ and $\beta=\frac{1}{2\gamma}$, we have
\be 
\mathbf{F}^{k}-\gamma\nabla h_s(\mathbf{F}^{k})&=&\Big((1-\gamma)\mathbf{I}-\gamma\lambda_f\tilde{\mathbf{L}}\Big)\mathbf{F}^{k}+\gamma \mathbf{X}_{trans}, \nonumber\\
&=&\gamma \mathbf{X}_{trans} + (1-\gamma)\tilde{\mathbf{A}}\mathbf{F}^{k},
\ee 
Therefore, we can summarize the proposed FMP as two phases, including propagation with skip connection (Step \textbf{\ding{182}}) and bias mitigation (Steps \textbf{\ding{183}}-\textbf{\ding{186}}). For bias mitigation, Step \textbf{\ding{183}} updates the aggregated node features for fairness objective; Steps \textbf{\ding{184}} and \textbf{\ding{185}} aim to learn and ``reshape" perturbation vector in probability space, respectively. Step \textbf{\ding{186}} explicitly mitigate the bias of node features based on gradient descent on primal variable. The mathematical formulation is given as follows:
\be 
\left\{
\begin{array}{ll}
\mathbf{X}_{agg}^{k+1}=\gamma \mathbf{X}_{trans} + (1-\gamma)\tilde{\mathbf{A}}\mathbf{X}^{k}, & \text{Step \textbf{\ding{182}}}\\
\bar{\mathbf{F}}^{k+1}=\mathbf{X}_{agg}^{k+1}-\gamma \frac{\partial \langle \mathbf{p}, \mathbf{u}^{k}\rangle}{\partial \mathbf{F}}\Big|_{\mathbf{F}^{k}}, & \text{Step \textbf{\ding{183}}}\\
\bar{\mathbf{u}}^{k+1}=\mathbf{u}^{k}+\beta \mathbf{\Delta_{s}} SF(\bar{\mathbf{F}}^{k+1}), & \text{Step \textbf{\ding{184}}}\\
\mathbf{u}^{k+1}=\min\Big(|\bar{\mathbf{u}}^{k+1}|, \lambda_{f} \Big)\cdot sign(\bar{\mathbf{u}}^{k+1}), & \text{Step \textbf{\ding{185}}}\\
\mathbf{F}^{k+1}=\mathbf{X}_{agg}^{k+1}-\gamma \frac{\partial \langle \mathbf{p}, \mathbf{u}^{k+1}\rangle}{\partial \mathbf{F}}\Big|_{\mathbf{F}^{k}}. & \text{Step \textbf{\ding{186}}}
\end{array}
\right. \nonumber
\ee 
where $\mathbf{X}_{agg}^{k+1}$ represents the node features with normal aggregation and skip connection with the transformed input $\mathbf{X}_{trans}$.

\section{More Discussions on Message Passing}\label{app:discussions}
In this section, we provide more discussion on the influence of $4$ graph data in-formation of graph:the number of nodes $n$, the edge density $\rho_d$, sensitive homophily coefficient $\epsilon_{sens}$, and sensitive group ratio $c$ for bias enhancement. 

\paragraph{On the sensitive homophily coefficient $\epsilon_{sens}$.} According to Lemma~\ref{lemma:statistical}, for large sensitive homophily coefficient $\epsilon_{sens}\rightarrow 1$, the inter-connect probability $q_{conn}\rightarrow 0$ and intra-connect probability $p_{conn}$ keeps the maximal value. In this case, 
based on Theorem~\ref{theo:enhance}, it is easy to obtain that $\nu_1=1$, $\nu_2=0$ and
the distance for the mean aggregated node representation will keep the same, i.e.,  $\tilde{\mathbf{\mu}}_1-\tilde{\mathbf{\mu}}_2=(\nu_1-\nu_2)(\mathbf{\mu}_1-\mathbf{\mu}_2)=\mathbf{\mu}_1-\mathbf{\mu}_2$. Additionally, the covariance will be diminished after aggregation since $\zeta_1$ and $\zeta_2$ are strictly larger than $1$. Therefore, for sufficient large sensitive homophily coefficient $\epsilon_{sens}$, the bias-enhance condition $(\nu_1-\nu_2)^2\min\{\zeta_1, \zeta_2\}>1$ holds. 

\paragraph{On the graph node number $n$.} For large node number $n$, the mean distance almostly keeps constant since $\nu_1\approx\frac{c p_{conn}}{(1-c)q_{conn}+c p_{conn}}$, $\nu_2\approx\frac{c q_{conn}}{(1-c)p_{conn}+c q_{conn}}$, and $\zeta_1, \zeta_2$ are almost proportional to node number $n$. Therefore, the bias-enhancement condition can be more easily satisfied and $\Delta Bias$ would be higher for large graph data. The intuition is that, given graph density ratio, large graph data represents higher average degree node. Hence, each aggregated node representation adopt more neighbor's information with the same sensitive attribute, and thus leads to lower covariance matrix value and higher representation bias.

\paragraph{On the graph connection density $\rho_d$.} Based on Lemma~\ref{lemma:statistical}, inter-connection probability and inter-connection probability are both proportional to graph connection density $\rho_d$. Therefore, $\nu_1$ and $\nu_2$ almost keep constant and the distance of mean node representation is constant as well. As for covariance matrix, message passing leads to more concentrated node representation since $\zeta_1$ and $\zeta_2$ are larger for higher graph connection density $\rho_d$. The rationale is similar with the graph node number: given node number, higher graph connection density $\rho_d$ means higher average node degree and each aggregated node representation adopt more neighbor's information with the same sensitive attribute.

\paragraph{On the sensitive attribute ration $c$.} Based on Lemma~\ref{lemma:statistical}, it is easy to obtain that, given graph connection density $\rho_d$ and graph node number $n$, the intra-connection probability $p_{conn}$ would be high, while low for inter-connection probability $q_{conn}$, if the balanced sensitive attribute. In other words,  intra-connection probability $p_{conn}$ (inter-connection probability $q_{conn}$) monotonically decreases (increases) with respect to $|c-\frac{1}{2}|$. 

\section{Proof of Theorem~\ref{theo:grad_comp}}
Before providing in-depth analysis on the gradient computation, we first introduce the softmax function derivative property in the following lemma:
\begin{lemma}\label{lemma:softmax}
For the softmax function with $N$-dimensional vector input $\mathbf{y}=SF(\mathbf{x}):\mathbb{R}^{1\times N} \longrightarrow\mathbb{R}^{1\times N}$, where $y_j=\frac{e^{\mathbf{x}_j}}{\sum_{k=1}^{N}e^{\mathbf{x}_k}}$ for $\forall j\in \{1,2,\cdots, N\}$, the derivative $N\times N$ Jocobian matrix is defined by $[\frac{\partial \mathbf{y}}{\partial \mathbf{x}}]_{ij}=\frac{\partial \mathbf{y}_i}{\partial \mathbf{x}_j}$. Additionally, Jocobian matrix satisfies $\frac{\partial \mathbf{y}}{\partial \mathbf{x}}=\mathbf{y} \mathbf{I}_{N}-\mathbf{y}^{\top}\mathbf{y}$, where $\mathbf{I}_{N}$ represents $N\times N$ identity matrix and $\top$ denotes the transpose operation for vector or matrix.
\begin{proof}
Considering the gradient for arbitrary $i= j$, according to quotient and chain rule of derivatives, we have
\be 
\frac{\partial \mathbf{y}_i}{\partial \mathbf{x}_j}=\frac{e^{\mathbf{x}_i}\sum_{k=1}^{N}e^{\mathbf{x}_k} -e^{\mathbf{x}_i+\mathbf{x}_j}}{\big(\sum_{k=1}^{N}e^{\mathbf{x}_k}\big)^2}=\frac{e^{\mathbf{x}_i}}{\sum_{k=1}^{N}e^{\mathbf{x}_k}} \cdot \frac{\sum_{k=1}^{N}e^{\mathbf{x}_k}-e^{\mathbf{x}_i}}{\sum_{k=1}^{N}e^{\mathbf{x}_k}}=\mathbf{y}_i (1-\mathbf{y}_j),
\ee 
Similarly, for arbitrary $i\neq j$, the gradient is given by
\be 
\frac{\partial \mathbf{y}_i}{\partial \mathbf{x}_j}=\frac{e^{\mathbf{x}_i}}{\sum_{k=1}^{N}e^{\mathbf{x}_k}} \cdot \frac{-e^{\mathbf{x}_i}}{\sum_{k=1}^{N}e^{\mathbf{x}_k}}=-\mathbf{y}_i\mathbf{y}_j.
\ee 
Combining these two cases, it is easy to verify the Jocobian matrix satisfies $\frac{\partial \mathbf{y}}{\partial \mathbf{x}}=\mathbf{y} \mathbf{I}_{N}-\mathbf{y}^{\top}\mathbf{y}$.
\end{proof}
\end{lemma}
Arming with the derivative property of softmax function, we further investigate the gradient $\frac{\partial \langle \mathbf{p}, \mathbf{u}\rangle}{\partial \mathbf{F}}$, where $\mathbf{p}=\mathbf{\Delta}_s SF(\mathbf{F})\in \mathbb{R}^{1\times d_{out}}$ and $SF(\cdot)$ and $\mathbf{u}\in \mathbb{R}^{1\times d_{out}}$ is independent with $\mathbf{F}\in \mathbb{R}^{n\times d_{out}}$. 

Considering softmax function $SF(\mathbf{x})\in\mathbb{R}^{n\times d}$ is row-wise adopted in node representation matrix, the gradient satisfies $\frac{\partial SF(\mathbf{F})_{i}}{\partial \mathbf{F}_j}=\mathbf{0}_{d_{out}\times d_{out}}$ for $i\neq j$. Note that the inner product $\langle \mathbf{p}, \mathbf{u}\rangle=\sum_{k=1}^{d_{out}} \mathbf{p}_k \mathbf{u}_k$, it is easy the obtain the gradient $[\frac{\partial \langle \mathbf{p}, \mathbf{u}\rangle}{\partial \mathbf{F}}]_{ij}=\sum_{k=1}^{d_{out}} \frac{\partial \mathbf{p}_k}{\partial \mathbf{F}_{ij}} \mathbf{u}_k$.

To simply the current notation, we denote $\mathbf{\tilde{F}}\dff SF(\mathbf{F})$. According to chain rule of derivative, we have
\be 
\frac{\partial \mathbf{p}_k}{\partial \mathbf{F}_{ij}} =
\sum_{t=1}^{d_{out}}\frac{\partial \mathbf{p}_k}{\partial \tilde{\mathbf{F}}_{tk}}\frac{\partial \tilde{\mathbf{F}}_{tk}}{\partial \mathbf{F}_{ij}}=
\sum_{t=1}^{d_{out}}\Delta_{\mathbf{s}, t}\frac{\partial \tilde{\mathbf{F}}_{tk}}{\partial \mathbf{F}_{ij}} \overset{(a)}{=}\Delta_{\mathbf{s}, i}\frac{\partial \tilde{\mathbf{F}}_{ik}}{\partial \mathbf{F}_{ij}}\overset{(b)}{=}\Delta_{\mathbf{s}, i}\tilde{\mathbf{F}}_{ik}[\delta_{kj}-\tilde{\mathbf{F}}_{ij}],
\ee 
where $\delta_{kj}$ is Dirac function (equals $1$ only if $k=j$, otherwise $0$;), equality (a) holds since softmax function is row-wise operation, and equality (b) is based on Lemma~\ref{lemma:softmax}. Furthermore, we can obtain the gradient of fairness objective w.r.t. node presentation as follows:
\be 
[\frac{\partial \langle \mathbf{p}, \mathbf{u}\rangle}{\partial \mathbf{F}}]_{ij}=\sum_{k=1}^{d_{out}} \frac{\partial \mathbf{p}_k}{\partial \mathbf{F}_{ij}} \mathbf{u}_k=\sum_{k=1}^{d_{out}}\Delta_{\mathbf{s}, i}\tilde{\mathbf{F}}_{ik}[\delta_{kj}-\tilde{\mathbf{F}}_{ij}]\mathbf{u}_k=\Delta_{\mathbf{s}, i}\tilde{\mathbf{F}}_{ij}\mathbf{u}_j-\Delta_{\mathbf{s}, i}\tilde{\mathbf{F}}_{ij}\sum_{k=1}^{d_{out}}\tilde{\mathbf{F}}_{ik}\mathbf{u}_k.
\ee 
Therefore, the matrix formulation is given by
\be 
\frac{\partial \langle \mathbf{p}, \mathbf{u}\rangle}{\partial \mathbf{F}}= \mathbf{U}_s\odot SF(\mathbf{F})-\text{Sum}_{1}(\mathbf{U}_s\odot SF(\mathbf{F}))SF(\mathbf{F}).
\ee 
where $\mathbf{U}_s\dff \Delta_s^{\top}\mathbf{u}\in\mathbb{R}^{n\times d_{out}}$ and $\text{Sum}_{1}(\cdot)$ represents the summation over column dimension with preserved matrix shape. Therefore, the computation complexity for gradient $\frac{\partial \langle \mathbf{p}, \mathbf{u}\rangle}{\partial \mathbf{F}}$ is $O(n d_{out})$.

\section{Proof of Proposition~\ref{prop:conjugate}}
We firstly show the conjugate function for general norm function $f(\mathbf{x})=\lambda ||\mathbf{x}||$, where $\mathbf{x}\in\mathbf{R}^{1\times d_{out}}$. The conjugate function of $f(\mathbf{x})$ satisfies 
\be 
f^{*}(\mathbf{y})=\left\{
\begin{array}{lr}
0, &||\mathbf{x}||_{*}\leq \lambda,\\
+\infty, & ||\mathbf{x}||_{*} > \lambda.
\end{array}
\right.
\ee 
where $||\mathbf{x}||_{*}$ is dual norm of the original norm $||\mathbf{x}||$, defined as $||\mathbf{y}||_{*}=\max\limits_{||\mathbf{x}||\leq 1}\mathbf{y}^{\top}\mathbf{x}$. Considering the conjugate function definition $f^{*}(\mathbf{y})=\max\limits_{\mathbf{x}}\mathbf{y}^{\top}\mathbf{x}-\lambda ||\mathbf{x}||$
the analysis can be divided as the following two cases:

\textbf{\ding{182}} If $||\mathbf{y}||_{*}\leq\lambda$, according to the definition of dual norm, we have $\mathbf{y}^{\top}\mathbf{x}\leq ||\mathbf{x}||||\mathbf{y}||_{*}\leq\lambda ||\mathbf{x}||$ for $\forall ||\mathbf{x}||$, where the equality holds if and only if $||\mathbf{x}||=0$. Hence, it is easy to obtain $f^{*}(\mathbf{y})=\max\limits_{\mathbf{x}}\mathbf{y}^{\top}\mathbf{x}-\lambda ||\mathbf{x}||=0$. 

\textbf{\ding{183}} If $||\mathbf{y}||_{*}>\lambda$, note that the dual norm $||\mathbf{y}||_{*}=\max\limits_{||\mathbf{x}||\leq 1}\mathbf{y}^{\top}\mathbf{x}> \lambda$, there exists variables $\hat{\mathbf{x}}$ so that $||\hat{\mathbf{x}}||\leq 1$ and $\hat{\mathbf{x}}^{\top}\mathbf{y}< \lambda$. Therefore, for any constant $t$, we have $f^{*}(\mathbf{y})\geq \mathbf{y}^{\top}(t\mathbf{x})-\lambda ||t\mathbf{x}||=t(\mathbf{y}^{\top}\mathbf{x}-\lambda ||\mathbf{x}||)\overset{t\rightarrow \infty}{\longrightarrow}\infty$.

Based on aforementioned two cases, it is easy to get the conjugate function for $l_1$ norm (the dual norm is $l_{\infty}$), i.e., the conjugate function for $h_f(\mathbf{x})=\lambda ||\mathbf{x}||_1$ is given by
\be 
h_{f}^{*}(\mathbf{y})=\left\{
\begin{array}{lr}
0, &||\mathbf{x}||_{\infty}\leq \lambda,\\
+\infty, & ||\mathbf{x}||_{\infty} > \lambda.
\end{array}
\right.
\ee 
Given the conjugate function $h_{f}^{*}(\cdot)$, we further investigate the proximal operators $\text{prox}_{h^{*}_{f}}$. Note that $\text{prox}_{h^{*}_{f}}(\mathbf{u})=\arg\min\limits_{\mathbf{y}}||\mathbf{y}-\mathbf{u}||_F^2+h^{*}_{f}(\mathbf{u})=\arg\min\limits_{||\mathbf{y}||_{\infty}\leq\lambda_{f}}||\mathbf{y}-\mathbf{u}||_F^2=\arg\min\limits_{\substack{\mathbf{y}_j\leq\lambda_{f} \\ \forall j\in[d_{out}]}}\sum_{j=1}^{d_{out}}|\mathbf{y}_j-\mathbf{u}_{j}|^2$, the proximal operator problem can be decomposed as element-wise sub-problem, i.e.,
\be 
\text{prox}_{h^{*}_{f}}(\mathbf{u})_{j}=\arg\min\limits_{\mathbf{y}_j\leq\lambda_{f}}|\mathbf{y}_j-\mathbf{u}_{j}|^2=sign(\mathbf{u}_j) \min(|\mathbf{u}_j|, \lambda_{f}) \nonumber
\ee  
which completes the proof.

\section{Dataset Statistics}\label{app:stat}
For fair comparison with previous work, we perform the node classification task on three real-world dataset, including Pokec-n, Pokec-z, and NBA.
The data statistical information on three real-world dataset is provided in Table~\ref{table:statistics}. It is seen that the sensitive homophily are even higher than label homophily coefficient among three real-world dataset, which validates that the real-world dataset are usually with large topology bias. 

\begin{table*}[h]
\begin{center}
\caption{Statistical Information on Datasets}
\label{table:statistics}
\scalebox{0.92}{
    \begin{tabular}{ c|c|c|c|c|c|c|c} 
    \toprule
    Dataset & $\#$ Nodes & \makecell*[c]{$\#$ Node Features} & $\#$ Edges & $\#$ Training Labels & $\#$ Training Sens & Label Homop & Sens Homop \\
    \hline
    Pokec-n &  66569 & 265 & 1034094 & 4398 & 500 & 73.23$\%$ & 95.30$\%$ \\
    \hline
    Pokec-z & 67796 & 276 & 1235916 & 5131 & 500 & 71.16$\%$ & 95.06$\%$ \\
    \hline
    NBA & 403 & 95 & 21242 & 156 & 246 & 39.22$\%$ & 72.37$\%$ \\
    \bottomrule
    \end{tabular}
}
\end{center}
\end{table*}

\section{More Discussion on GNNs as Graph Signal Denoising}\label{app:denoise}
In this section, we provide more examples to show many existing GNNs can be interpreted as graph signal denoising problem, including GAT and APPNP. 
\paragraph{GAT.} Feature aggregation in GAT applies the normalized attention coefficient to compute a linear combination of neighbor's features as $\mathbf{X}_{agg, i}=\sum_{j\in\mathcal{N}(i)}\alpha_{ij}\mathbf{X}_{trans, j}$, where $\alpha_{ij}=softmax_j(e_{ij})$, $e_{ij}=\text{LeakyReLU}(\mathbf{X}_{trans, i}^{\top}\mathbf{w}_i+\mathbf{X}_{trans, j}^{\top}\mathbf{w}_j)$, and $\mathbf{w}_i$ and $\mathbf{w}_j$ are learnable column vectors. Prior study \cite{ma2021unified} demonstrates that one-step gradient descent with adaptive stepsize $\frac{1}{\sum_{j\in\tilde{\mathcal{N}}(i)}(c_i+c_j)}$ for the following objective problem:
\be 
\min\limits_{\mathbf{F}}\sum_{i\in\mathcal{V}}||\mathbf{F}_i-\mathbf{X}_{trans, i}||^2_F + \frac{1}{2}\sum_{i\in\mathcal{V}}c_i\sum_{j\in\tilde{\mathcal{N}}(i)}||\mathbf{F}_i-\mathbf{F}_j||_F^2. \nonumber
\ee 
is actually an attention-based feature aggregation, which equivalent to GAT if $c_i+c_j$ is equivalent to $e_{ij}$, where $c_i$ is a node-dependent coefficient that measures the local smoothness.

\paragraph{PPNP / APPNP.} Feature aggregation in PPNP and APPNP adopt the aggregation rules as $\mathbf{X}_{agg}=\alpha\Big(\mathbf{I}-(1-\alpha)\tilde{\mathbf{A}}\Big)^{-1}\mathbf{X}_{trans}$ and $\mathbf{X}_{agg}^{k+1}=(1-\alpha)\tilde{\mathbf{A}}\mathbf{X}_{agg}^{k}+\alpha\mathbf{X}_{trans}$. It is shown that they are equivalent to the exact solution and one gradient descent step with stepsize $\frac{\alpha}{2}$ to minimize the following objective problem:
\be
\min\limits_{\mathbf{F}}||\mathbf{F}-\mathbf{X}_{trans}||_F^2+(\frac{1}{\alpha}-1)tr\Big(\mathbf{F}^{\top}(\mathbf{I}-\tilde{\mathbf{A}})\mathbf{F}\Big). \nonumber
\ee

\section{Related Works} \label{app:related}
\paragraph{Graph neural networks.} GNNs generalizing neural networks for graph data have already been shown great success for various real-world applications. There are two streams in GNNs model design, i.e., spectral-based and spatial-based. Spectral-based GNNs provide graph convolution definition based on graph theory, which is utilized in GNN layers together with feature transformation \cite{bruna2013spectral,defferrard2016convolutional,henaff2015deep}. Graph convolutional networks (GCN) \cite{kipf2017semi} simplifies spectral-based GNN model into spatial aggregation scheme. Since then, many spatial-based GNNs variant are developed to update node representation via aggregating its neighbors' information, including graph attention network (GAT) \cite{velivckovic2018graph},  GraphSAGE \cite{hamilton2017inductive}, SGC \cite{wu2019simplifying}, APPNP \cite{klicpera2019predict}, et al \cite{gao2018large,monti2017geometric}. Graph signal denoising is another perspective to understand GNNs. Recently, there are several works show that GCN is equivalent to the first-order approximation for graph denoising with Laplacian regularization \cite{henaff2015deep,zhao2019pairnorm}. The unified optimization framework are provided to unify many existing message passing schemes \cite{ma2021unified,zhu2021interpreting}. 

\paragraph{Fairness-aware learning on graphs.} Many works have been developed to achieve fairness in machine learning community \cite{chuang2020fair,zhang2018mitigating,du2021fairness,yurochkin2020sensei,creager2019flexibly,feldman2015certifying}. A pilot study on fair node representation learning is developed based on random walk \cite{rahman2019fairwalk}. Additionally, adversarial debiasing is adopt to learn fair prediction or node representation so that the well-trained adversary can not predict the sensitvie attribute based on node representation or prediction \cite{dai2021say,bose2019compositional,fisher2020debiasing}. A Bayesian approach is developed to learn fair node representation via encoding sensitive information in prior distribution in \cite{buyl2020debayes}. Work \cite{ma2021subgroup} develops a PAC-Bayesian analysis to connect subgroup generalziation with accuracy parity. \cite{laclau2021all,li2021dyadic} aim to mitigate prediction bias for link prediction. Fairness-aware graph contrastive learning are proposed in \cite{agarwal2021towards,kose2021fairness}. However, aforementioned works ignore the role of graph topology information in terms of fairness. In this work, we theoretically and experimentally reveal that many GNNs aggregation schemes boost node representation bias under topology bias. Furthermore, we develop a efficient and transparent fair message passing scheme with theoretical guarantee utilizing useful information of neighbors and debiasing node representation simultaneously. 

\section{More Experimental Results}\label{app:more_exp}
\subsection{More Synthetic Experimental Results} 
In this subsection, we provide more experimental results on \textbf{different} covariance matrix. Although our theory is only derived for the same covariance matrix, we still observe similar results for the case of \textbf{different} covariance matrix.
For \textbf{node attribute generation}, we generate node attribute with Gaussian distribution $\mathcal{N}(\mathbf{\mu}_1, \mathbf{\Sigma})$ and $\mathcal{N}(\mathbf{\mu}_2, \mathbf{\Sigma})$ for node with binary sensitive attribute, respectively, where $\mathbf{\mu}_1=[0,1]$, $\mathbf{\mu}_1=[1,0]$ and $\mathbf{\Sigma}=\left[ {\begin{array}{cc}
   1 & 0 \\
   0 & 2 \\
  \end{array} } \right]$. 
We adopt the same evaluation metric and adjacency matrix generation scheme in Section~\ref{subsect:syn}

Figure\ref{fig:dp_diff2} shows the DP difference during message passing with respect to sensitive homophily coefficient for different intial covariance matrix. We observes that higher sensitive homophily coefficient generally leads to larger bias enhancement. Additionally, higher graph connection density $\rho_d$, larger node number $n$, and balanced sensitive attribute ratio $c$ corresponds higher bias enhancement, which is consistent to our theoretical analysis in Theorem~\ref{theo:enhance}. 

\begin{figure}[t]
\centering
\includegraphics[width=0.99\linewidth]{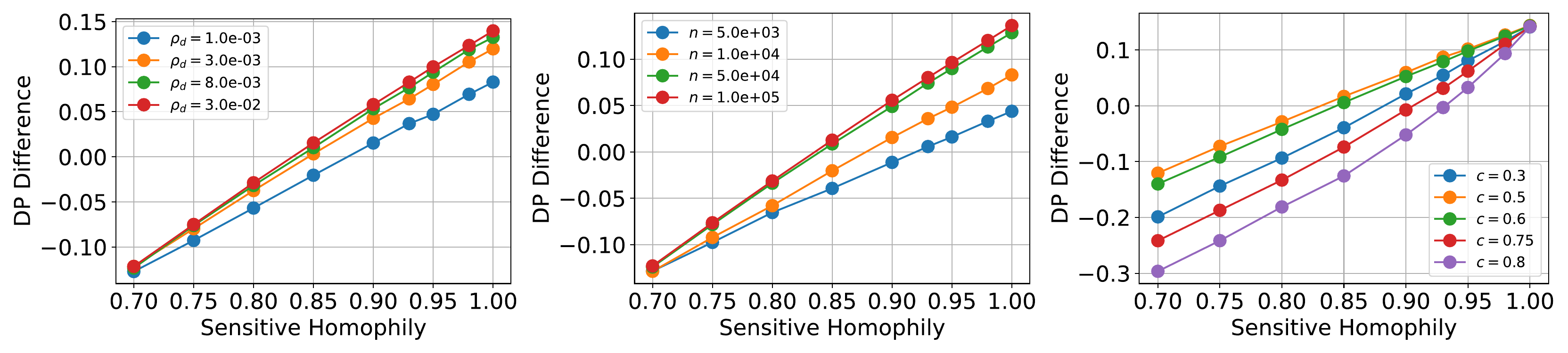}
\caption{The difference of demographic parity for message passing with different initial covariance matrix. \textbf{Left:} DP difference for different graph connection density $\rho_d$ with senstive attribute ratio $c=0.5$ and number of nodes $n=10^{4}$; \textbf{Middle:} DP difference for different number of nodes $n$ with senstive attribute ratio $c=0.5$ and graph connection density $\rho_d=10^{-3}$; \textbf{Right:} DP difference for different senstive attribute ratio $c$ with graph connection density $\rho_d=10^{-3}$ and number of nodes $n=10^{4}$.}
\label{fig:dp_diff2}
\end{figure}

\subsection{Hyperparameter Study} 
We provide hyperparameter study for further investigation on fairness and smoothness hyperparmeter on prediction and fairness performance on three datasets. Specifically, we tune hyperparameters as $\lambda_f=\{0.0, 5.0, 10.0, 15.0, 20.0, 30.0, 100.0, 1000.0\}$ and $\lambda_s=\{0.0, 0.1, 0.5, 1.0, 3.0, 5.0, 10.0, 15.0, 20.0\}$. From the results in Figure~\ref{fig:hyper}, we can make the following observations:
\begin{itemize}
    \item The accuracy and demographic parity are extremely sensitive to smoothness hyperparameter. It is seen that, for Pokec-n and Pokec-z datasets (NBA), larger smoothness hyperparameter usually leads to higher (lower) accuracy with higher prediction bias. The rationale is that, only for graph data with high label homophily coefficient, GCN-like aggregation with skip connection is beneficial. Otherwise, the neighbor's node representation with different label will mislead representation update.
    \item The appropriate fairness hyperparameter leads to better fairness and prediction performance tradeoff. The reason is that fairness hyperparameter determinates the perturbation vector update step size in probability space. Only appropriate step size can lead to better perturbation vector update.
\end{itemize}

\begin{figure}[t]
\centering
\includegraphics[width=0.99\linewidth]{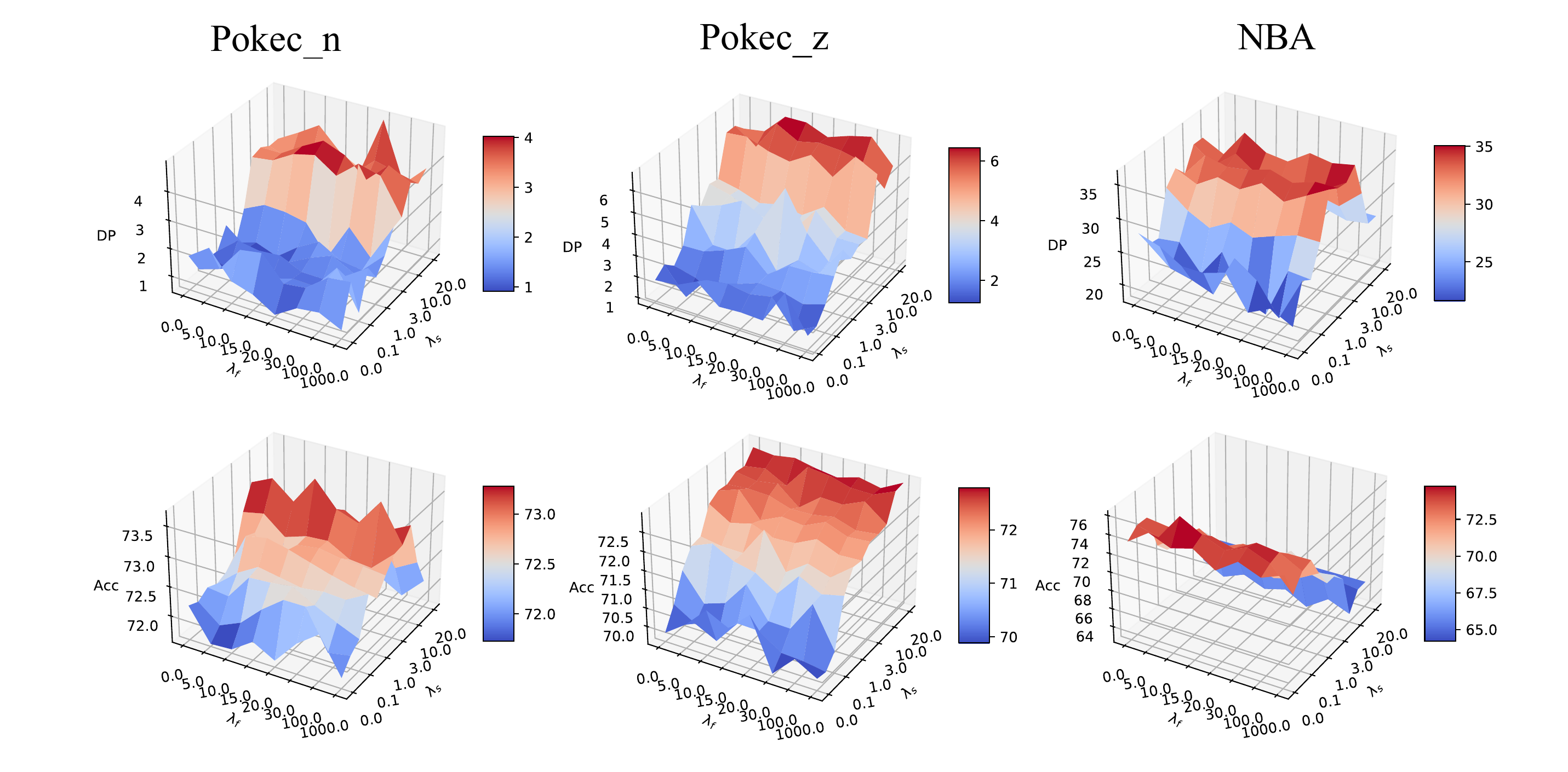}

\caption{Hyperparameter study on fairness and smoothness hyperparameter for demographic parity and Accuracy.}
\label{fig:hyper}
\end{figure}

\subsection{Running Time Comparison} 
We provide running time comparison in Table~\ref{fig:runtime} for our proposed FMP and other baselines, including vanilla, regularization and adversarial debiasing on many backbones (MLP, GCN, GAT, SGC, and APPNP). To achieve fair comparison, we adopt the same Adam optimizer with $200$ epochs with $5$ running times. 
We list several observations as follows:
\begin{itemize}
    \item The running time of proposed FMP is very efficient for large-scale dataset. Specifically, for vanilla method, the running time of FMP is higher than most lighten backbone MLP with $46.97\%$ and $15.03\%$ time overhead on Pokec-n and Poken-z dataset, respectively. Compared with the most time-consumption APPNP, the running time of FMP is lower with $64.07\%$ and $41.45\%$ time overhead on Pokec-n and Poken-z dataset, respectively.
    \item The regularization method achieves almost the same running time compared with vanilla method on all backbones. For example, GCN with regularization encompasses higher running time with $6.41\%$ time overhead compared with vanilla method. Adversarial debiasing is extremely time-consuming. For example, GCN with adversarial debiasing encompasses higher running time with $88.58\%$ time overhead compared with vanilla method.
\end{itemize}

\begin{figure*}[t]
\centering
\includegraphics[width=0.99\linewidth]{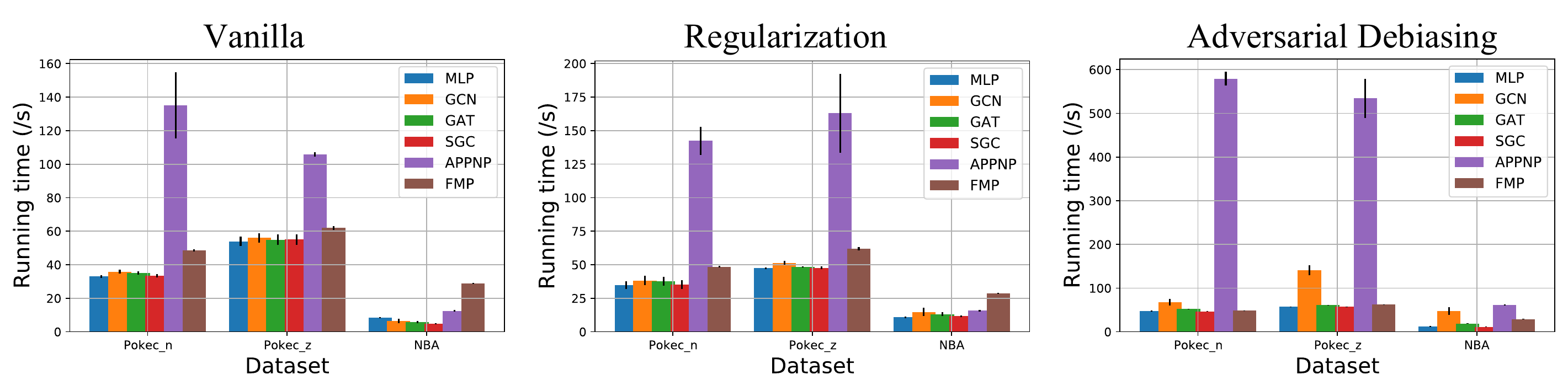}

\caption{The running time comparison.}
\label{fig:runtime}
\end{figure*}

\end{document}